\documentclass{article} 
\usepackage{iclr2020_conference,times}

\usepackage{hyperref}
\usepackage{url}

\usepackage[utf8]{inputenc} 
\usepackage[T1]{fontenc}    
\usepackage{hyperref}       
\usepackage{url}            
\usepackage{booktabs}       
\usepackage{amsfonts}       
\usepackage{nicefrac}       
\usepackage{microtype}      

\usepackage{bm}
\usepackage{color}
\usepackage{csquotes}
\usepackage{amssymb}
\usepackage{amsmath}
\usepackage{latexsym}
\usepackage{algpseudocode}
\usepackage{algorithm}

\usepackage{xspace}
\usepackage{soul}
\usepackage{dsfont}
\usepackage{stmaryrd}
\usepackage{todonotes}
\usepackage{dirtytalk}
\usepackage{pbox}
\usepackage{paralist}
\usepackage{cprotect}

\usepackage{tikz}
\usepackage{pgfplots}
\pgfplotsset{compat=1.9}
\usepackage{verbatim}
\usepackage{textcomp}
\usetikzlibrary{shapes,arrows,positioning}
\usepackage[normalem]{ulem}
\usetikzlibrary{calc}

\usepackage{mathtools}
\usepackage{etextools}
\DeclareMathAlphabet{\mathalphm}{OMS}{cmsy}{m}{n}
\DeclareMathAlphabet{\mathalphb}{OMS}{cmsy}{b}{n}

\newcommand{\R}{\mathbb{R} \xspace}

\newcommand{\E}{\mathop{\mathbb{E} \xspace}}

\newcommand{\norm}[1]{\left\lVert#1\right\rVert}

\newcommand{\pluseq}{\mathrel{+}=}
\newcommand{\diveq}{\mathrel{/}=}

\title{Min-max Entropy for Weakly Supervised Pointwise Localization}

\author{Soufiane Belharbi{\normalfont \textsuperscript{1}}\;, Jérôme Rony{\normalfont \textsuperscript{1}}, Jose Dolz{\normalfont \textsuperscript{1}}, Ismail Ben Ayed{\normalfont \textsuperscript{1}},
  {\bf Luke McCaffrey}{\normalfont \textsuperscript{2}}, \& {\bf Eric Granger}{\normalfont \textsuperscript{1}}  \\
{\normalfont \textsuperscript{1}} \'Ecole de technologie sup\'erieure, Universit\'e du Qu\'ebec, Montreal, Canada\\
{\normalfont \textsuperscript{2}} Rosalind and Morris Goodman Cancer Research Centre,   Dept. of Oncology, McGill University, Montreal, Canada\\
\texttt{\{soufiane.belharbi.1,jerome.rony.1\}@etsmtl.net} \\
  \texttt{\{jose.dolz,ismail.benayed,eric.granger\}@etsmtl.ca} \\
  \texttt{luke.mccaffrey@mcgill.ca}
}

\iclrfinalcopy 
\begin{document}

\maketitle

\begin{abstract}
Pointwise localization allows more precise localization and accurate interpretability, compared to bounding box, in applications where objects are highly unstructured such as in medical domain. In this work, we focus on  weakly supervised localization (WSL) where a model is trained to classify an image and localize regions of interest at pixel-level using only global image annotation. Typical convolutional attentions maps are prune to high false positive regions. To alleviate this issue, we propose a new deep learning method for WSL, composed of a localizer and a classifier, where the localizer is constrained to determine relevant and irrelevant regions using conditional entropy (CE) with the aim to reduce false positive regions.
Experimental results on a public medical dataset and two natural datasets, using Dice index, show that, compared to state of the art WSL methods, our proposal can provide significant improvements in terms of image-level classification and pixel-level localization (low false positive) with robustness to overfitting. A public reproducible PyTorch implementation is provided\footnote{\href{https://github.com/sbelharbi/wsol-min-max-entropy-interpretability}{https://github.com/sbelharbi/wsol-min-max-entropy-interpretability}}.
\end{abstract}

\section{Introduction}
\label{sec:introduction}
Pointwise localization is an important task for image understanding, as it provides crucial clues to challenging  visual recognition problems, such as semantic segmentation, besides being an essential and precise visual interpretability tool. Deep learning methods, and particularly convolutional neural networks (CNNs), are driving recent progress in these tasks. Nevertheless, despite their remarkable performance, their training requires large amounts of labeled data, which is time consuming and prone to observer variability. To overcome this limitation, weakly supervised learning (WSL) has emerged recently as a surrogate for extensive annotations of training data \citep{zhou2017brief}. WSL involves scenarios where training is performed with inexact or uncertain supervision. In the context of
pointwise localization or semantic segmentation, weak supervision typically comes in the form of image level tags \citep{KERVADEC201988,kim2017two,pathak2015constrained,teh2016attention,wei2017object}, scribbles \citep{Lin2016,ncloss:cvpr18} or bounding boxes \citep{Khoreva2017}.

Current state-of-the-art WSL methods rely heavily on pixelwise activation maps produced by a CNN classifier at the image level, thereby localizing regions of interest \citep{zhou2016learning}. Furthermore, this can be  used as an \emph{interpretation} of the model's decision \citep{Zhang2018VisualInterp}. The recent literature abounds of WSL works that relax the need of dense and prohibitively time consuming pixel-level annotations \citep{rony2019weak-loc-histo-survey}.
Bottom-up methods rely on the input signal to locate regions of interest, including spatial pooling techniques over activation maps \citep{durand2017wildcat,oquab2015object,sun2016pronet,zhang2018adversarial,zhou2016learning}, multi-instance learning \citep{ilse2018attention} and attend-and-erase based methods \citep{kim2017two,LiWPE018CVPR,pathak2015constrained,SinghL17,wei2017object}. While these methods provide pointwise localization, the models
in \citep{bilen2016weakly,kantorov2016contextlocnet,shen2018generative,tang2017multiple,wan2018min} predict a bounding box instead, i.e., perform weakly supervised object detection. Inspired by human visual attention, top-down methods rely on the input signal and a selective backward signal to determine the corresponding region of interest. This includes special feedback layers \citep{cao2015look}, backpropagation error \citep{zhang2018top} and Grad-CAM \citep{ChattopadhyaySH18wacv,selvaraju2017grad}.

\begin{figure}[h!]
  \center
\includegraphics[width=.8\linewidth]{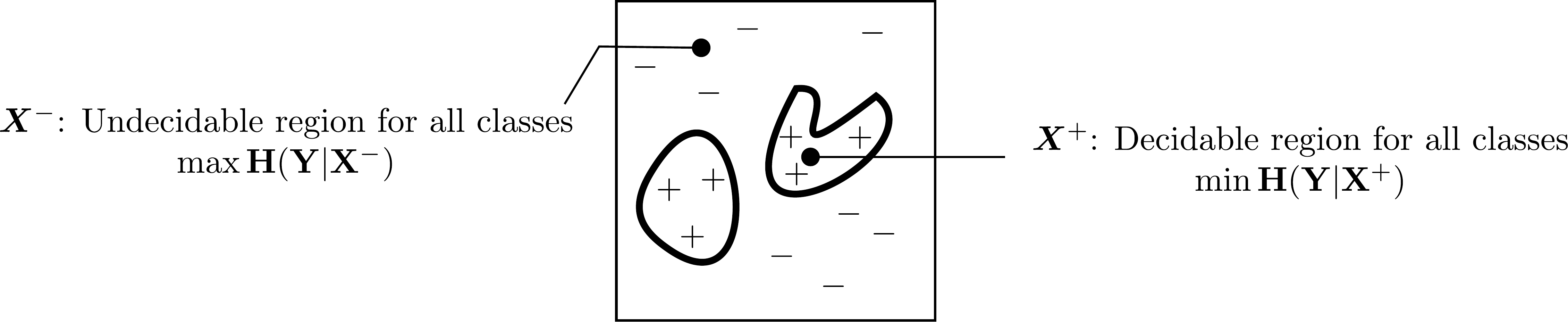}\\
\vspace{2mm}
\includegraphics[width=.8\linewidth]{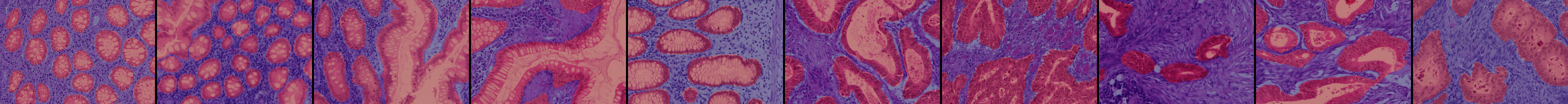}
\caption{\textbf{Top row}: Intuition of our proposal. A decidable region covers all the discriminative parts, while an undecidable region covers all the non-discriminative parts. (See Sec.\ref{sec:proposedmethod} for notation.) \textbf{Bottom row}: Example of test samples from different classes of the GlaS dataset, where the annotated glands are regions of interest, and the remaining tissue is noise/background. Note the glands' different shapes, size, context, and multiple instances aspect.}
\label{fig:fig-0}
\end{figure}

In many applications, such as in medical imaging, region localization may require high precision such as cells, boundaries, and organs localization; regions that have an unstructured shape, and different scale that a bounding box may not be able to localize precisely. In such cases, a pointwise localization can be more suitable. The illustrative example in Fig.\ref{fig:fig-0} (bottom row) shows a typical case where using a bounding box to localize the glands is clearly problematic. This motivates us to consider predicting a mask instead of a bounding box. Consequently, our latter choice of evaluation datasets is constrained by the availability of both global image annotation for training and pixel-level annotation for evaluation. In this work, we focus on the case where there is one object of interest in the image.

Often, within an agnostic-class setup, input image contains the object of interest among other irrelevant parts (noise,  background). Most the aforementioned WSL methods do not consider such prior, and feed the entire image to the model. In such scenario, \citep{wan2018min} argue that there is an \emph{inconsistency} between the classification loss and the task of WSL; and that typically the optimization may reach sub-optimal solutions with considerable randomness in them, leading to high false positive localization. False positive localization is aggravated when a class appears in different and random shape/structure, or may have relatively similar texture/color to the irrelevant parts driving the model to confuse between both parts. False positive regions can be problematic in critical domains such as medical applications where interpretability plays a central role in trusting and understanding an algorithm's prediction. To address this important issue, and motivated by the importance of using prior knowledge in learning to alleviate overfitting when training using few samples \citep{sbelharbiarxivsep2017,krupka2007incorporating,mitchell1980need,yu2007incorporating}, we propose to use the aforementioned prior in order to favorite models with low false positive localization. To this end, we constrain the model to learn to localize both relevant and irrelevant regions \textit{simultaneously} in an end-to-end manner within a WSL scenario, where only image-level labels are used for training. We model the relevant (discriminative) regions as the \emph{complement} of the irrelevant (non-discriminative) regions (Fig.\ref{fig:fig-0}). Our model is composed of two sub-models: \begin{inparaenum}[(1)]
\item a \emph{localizer} that aims to localize both types of regions by predicting a \emph{latent mask},
\item and a \emph{classifier} that aims to classify the visible content of the input image through the latent mask.
\end{inparaenum}
The \emph{localizer} is driven through CE \citep{coverentropy2006} to simultaneously identify
\begin{inparaenum}[(1)]
\item relevant regions where the \emph{classifier} has high confidence with respect to the image label,
\item and irrelevant regions where the classifier is being unable to decide which image label to assign.
\end{inparaenum}
This modeling allows the discriminative regions to pop out and be used to assign the corresponding image label, while suppressing non-discriminative areas, leading to more reliable predictions. In order to localize complete discriminative regions, we extend our proposal by training the localizer to recursively erase discriminative parts during \emph{training} only. To this end, we propose a consistent recursive erasing algorithm that we incorporate within the backpropagation. At each recursion, and within the backpropagation, the algorithm localizes the most discriminative region; stores it; then erases it from the input image. At the end of the final recursion, the model has gathered a large extent of the object of interest that is fed next to the classifier. Thus, our model is driven to localize complete relevant regions while discarding irrelevant regions, resulting in more reliable region localization. Moreover, since the discriminative parts are allowed to be extended over different instances, our proposal handles multi-instances intrinsically.

The main contribution of this paper is a new deep learning framework for WSL at pixel level. The framework is composed of two sequential sub-networks where the first one localizes regions of interest, whereas the second classifies them. Based on CE, the end-to-end training of the framework allows to incorporate prior knowledge that, an image is more likely to contain relevant and irrelevant regions. Throughout the CE measured at the classifier level, the localizer is driven to localize relevant regions (with low CE) and irrelevant regions (with high CE). Such localization is achieved with the main goal of providing a more interpretable and reliable regions of interest with low false positive localization. This paper also contributes a consistent recursive erasing algorithm that is incorporated within backpropagation, along with a practical implementation in order to obtain complete discriminative regions. Finally, we conduct an extensive series of experiments on three public image datasets (medical and natural), where the results show the effectiveness of the proposed approach in terms of pointwise localization (measured with Dice index) while maintaining competitive accuracy for image-level classification.

\section{Background on WSL}
\label{sec:relatedwork}
In this section, we briefly review state of the art of WSL methods, divided into two main categories, aiming at pointwise localization of regions of interest using only image-level labels as supervision.
(1) Fully convolutional networks with spatial pooling have shown to be effective to obtain localization of discriminative regions \citep{durand2017wildcat,oquab2015object,sun2016pronet,zhang2018adversarial,zhou2016learning}. Multi-instance learning methods have been used within an attention framework to localize regions of interest \citep{ilse2018attention}. \citep{SinghL17} propose to hide randomly large patches in training image in order to force the network to seek other discriminative regions to recover large part of the object of interest, since neural networks often provide small and most discriminative regions of object of interest \citep{kim2017two,SinghL17,zhou2016learning}. \citep{wei2017object} use the attention map of a trained network to erase the most discriminative part of the original image. \citep{kim2017two} use two-phase learning stage where the attention maps of two networks are combined to obtain a complete region of the object. \citep{LiWPE018CVPR} propose a two-stage approach where the first network classifies the image, and provides an attention map of the most discriminative parts. Such attention is used to erase the corresponding parts over the input image, then feed the resulting erased image to a second network to make sure that there is no discriminative parts left.
(2) Inspired by the human visual attention, top-down methods were proposed. In \citep{Simonyan14a,DB15a,zeiler2014ECCV}, backpropagation error is used in order to visualize saliency maps over the image for the predicted class. In \citep{cao2015look}, an attention map is built to identify the class relevant regions using feedback layer. \citep{zhang2018top} propose Excitation backprop that allows to pass along top-down signals downwards in the network hierarchy through a probabilistic framework. Grad-CAM \citep{selvaraju2017grad} generalize CAM \citep{zhou2016learning} using the derivative of the class scores with respect to each location on the feature maps; it has been furthermore generalized in \citep{ChattopadhyaySH18wacv}. In practice, top-down methods are considered as visual explanatory tools, and they can be overwhelming in term of computation and memory usage even during inference.

While the aforementioned approaches have shown great success mostly with natural images, they still lack a mechanism for modeling what is relevant and irrelevant within an image which is important to reduce false positive localization. This is crucial for determining the reliability of the regions of interest. Erase-based methods \citep{kim2017two,LiWPE018CVPR,pathak2015constrained,SinghL17,wei2017object} follow such concept where the non-discriminative parts are suppressed through constraints, allowing only the discriminative ones to emerge. Explicitly modeling negative evidence within the model has shown to be effective in WSL \citep{Azizpour2015SpotlightTN,durand2017wildcat,durand2016weldon,PariziVZF14}.

Our proposal is related to \citep{Behpour2019,wan2018min} in using entropy-measure to explore the input image. However, while \citep{wan2018min} defines an entropy over the bounding boxes' position to minimize its variance, we define a CE over the classifier to be low over discriminative regions, while being high over non-discriminative ones. Our recursive erasing algorithm follows general erasing and mining techniques \citep{kim2017two,LiWPE018CVPR,SinghL17,wan2018min,wei2017object}, but places more emphasis on mining consistent regions, and being performed on the fly during backpropagation. For instance, compared to \citep{wan2018min}, our algorithm attempts to expand regions of interest, accumulate consistent regions while erasing, provide automatic mechanism to stop erasing over samples independently from each other. However \citep{wan2018min} aims to locate multiple instances without erasing, and use manual/empirical threshold for assigning confidence to boxes. Our proposal can be seen as a \emph{guided} dropout \citep{srivastava14a}. While standard dropout is applied over a given input image to \emph{randomly} zero out pixels, our proposed approach \emph{seeks} to zero out irrelevant pixels and keep only the discriminative ones that support the image label. From this perspective, our proposal mimics a \emph{discriminative gate} that inhibits irrelevant and noisy regions while allowing only informative and discriminative regions to pass through the gate.

\section{The min-max entropy framework for WSL}
\label{sec:proposedmethod}
\textbf{Notations and definitions:}
Let us consider a set of training samples ${\mathbb{D} = \{(\bm{X}_i, y_i)\}_{i=1}^n}$ where ${\bm{X}_i}$ is an input image with depth $d$, height $h$, and width $w$; a realization of the discrete random variable ${\mathbf{X}}$ with support set ${\mathcal{X}}$; ${y_i}$ is the image-level label (i.e., image class), a realization of the discrete random variable ${\mathbf{y}}$ with support set ${\mathcal{Y} = \{1, \cdots, c\}}$. We define a \emph{decidable} region\footnote{In this context, the notion of \emph{region} indicates one pixel.} of an image as any informative part of the image that allows predicting the image label. An \emph{undecidable} region is any noisy, uninformative, and irrelevant part of the image that does not provide any indication nor support for the image class. To model such definitions, we consider a binary mask ${\bm{M}^+ \in \{0, 1\}^{h \times w}}$ where a location $(r, z)$ with value $1$ indicates a decidable region, otherwise it is an undecidable region. We model the decidability of a given location $(r, z)$ with a binary random variable ${\mathbf{M}}$. Its realization is ${\bm{m}}$, and its conditional probability  ${p_{\mathbf{m}}}$ over the input image is defined as follows,
\begin{equation}
  \label{eq:eq0}
  p_{\mathbf{M}}(\mathbf{m} = 1| \bm{X}, (r, z)) =
  \begin{cases}
    1 & \text{if } \bm{X}(r, z) \text{ is a decidable region}\; , \\
    0 & \text{otherwise.}
  \end{cases}
\end{equation}
We note ${\bm{M}^- \in \{0, 1\}^{h \times w} = \bm{U} - \bm{M}^+}$  a binary mask indicating the undecidable region, where ${\bm{U} = \{1\}^{h \times w}}$. We consider the undecidable region as the \emph{complement} of the decidable one. We can write: ${\norm{\bm{M}^+}_0 + \norm{\bm{M}^-}_0 = h \times w}$, where ${\norm{\cdot}_0}$ is the ${l_0}$ norm. Following such definitions, an input image ${\bm{X}}$ can be decomposed into two images as
$
{\bm{X} = \bm{X} \odot \bm{M}^+ + \bm{X} \odot \bm{M}^-}$,
where ${(\cdot \odot \cdot)}$ is the Hadamard product. We note ${\bm{X}^+ = \bm{X} \odot \bm{M}^+}$, and ${\bm{X}^- = \bm{X} \odot \bm{M}^-}$. ${\bm{X}^+}$ inherits the image-level label of ${\bm{X}}$. We can write the pair ${(\bm{X}^+_i, y_i)}$ in the same way as ${(\bm{X}_i, y_i)}$. We note by ${\bm{R}^+_i}$, and ${\bm{R}^-_i}$ as the respective approximation of ${\bm{M}^+_i}$, and ${\bm{M}^-_i}$. We are interested in modeling the true conditional distribution ${p(\mathbf{Y} | \mathbf{X})}$ where ${p(\mathbf{Y} = y_i | \mathbf{X} = \bm{X}_i) = 1}$. ${\hat{p}(\mathbf{Y} | \mathbf{X})}$ is its estimate. Following the previous discussion, predicting the image label depends only on the decidable region, i.e., ${\bm{X}^+}$. Thus, knowing ${\bm{X}^-}$ does not add any knowledge to the prediction, since ${\bm{X}^-}$ does not contain any information about the image label. This leads to:
$
{p(\mathbf{Y} | \mathbf{X} = \bm{X}) = p(\mathbf{Y} | \mathbf{X} = \bm{X}^+)}$. As a consequence, the image label is conditionally independent of ${\bm{X}^-}$ provided  ${\bm{X}^+}$ \citep{Kollergraphical2009}: ${p \models \mathbf{Y} \perp \mathbf{X}^- | \mathbf{X}^+}$, where ${\mathbf{X}^+, \mathbf{X}^-}$ are the random variables modeling the decidable and the undecidable regions, respectively. In the following, we provide more details on how to exploit such conditional independence property in order to estimate ${\bm{R}^+}$ and ${\bm{R}^-}$.

\begin{figure}[h!]
  \center
\includegraphics[width=.95\linewidth]{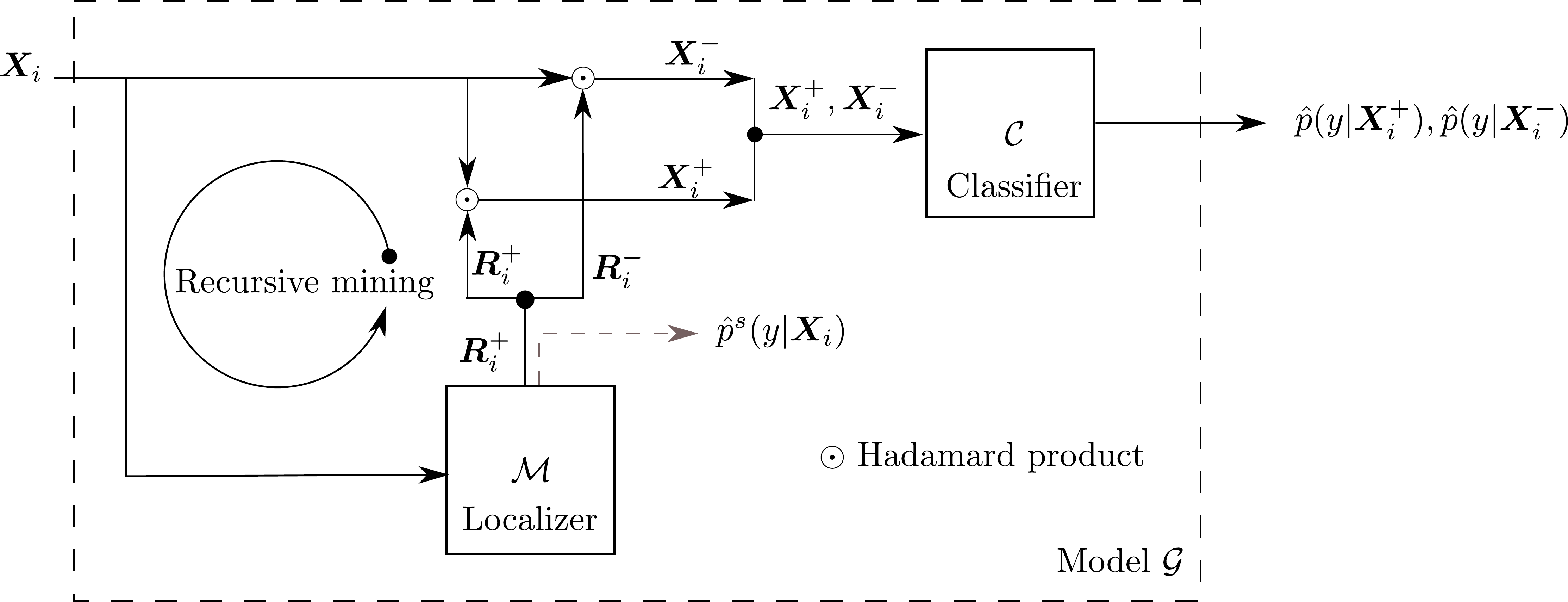}
\caption{Our proposed method. The recursive mining is done only during training (See Sec.\ref{sub:recursive-erasing-sup-mat}).}
\label{fig:fig-4}
\end{figure}

\textbf{Min-max entropy:}
We consider modeling the uncertainty of the model prediction over decidable, or undecidable regions using conditional entropy (CE). Let us consider the CE of ${\mathbf{Y} | \mathbf{X} = \bm{X}^+}$, denoted ${\mathbf{H}(\mathbf{Y} | \mathbf{X} = \bm{X}^+)}$ and computed as \citep{coverentropy2006},
\begin{equation}
  \label{eq:eq3}
  \mathbf{H}(\mathbf{Y} |  \mathbf{X} = \bm{X}^+) = - \sum_{y \in \mathcal{Y}} \hat{p}(\mathbf{Y} |  \mathbf{X} = \bm{X}^+) \; \log \hat{p}(\mathbf{Y} |  \mathbf{X} = \bm{X}^+) \; .
\end{equation}
Since the model is required to be certain about its prediction over ${\bm{X}^+}$, we constrain the model to have low entropy over ${\bm{X}^+}$. Eq.\ref{eq:eq3} reaches its minimum when the probability of one of the classes is certain, i.e., ${\hat{p}(\mathbf{Y}=y |  \mathbf{X} = \bm{X}^+) = 1}$ \citep{coverentropy2006}. Instead of directly minimizing Eq.\ref{eq:eq3}, and in order to ensure that the model predicts the correct image label, we cast a supervised learning problem using the cross-entropy between $p$ and ${\hat{p}}$ using the image-level label of ${\bm{X}}$ as a supervision,
\begin{align}
  \mathbf{H}(p_i, \hat{p}_i)^+ &= - \sum_{y \in \mathcal{Y}}  p(\mathbf{Y} = y |  \mathbf{X} = \bm{X}^+_i) \; \log \hat{p}(\mathbf{Y} = y |  \mathbf{X} = \bm{X}^+_i)
  = - \log \hat{p}(y_i | \bm{X}^+_i)  \label{eq:eq5}\; .
\end{align}
Eq.\ref{eq:eq5} reaches its minimum at the same conditions as Eq.\ref{eq:eq3} with the true image label as a prediction. We note that Eq.\ref{eq:eq5} is the negative log-likelihood of the sample ${(\bm{X}_i, y_i)}$. In the case of ${\bm{X}^-}$, we consider the CE of ${\mathbf{Y} |  \mathbf{X} =  \bm{X}}^-$, denoted ${\mathbf{H}(\mathbf{Y} |  \mathbf{X} = \bm{X}^-)}$ and computed as,
\begin{equation}
  \label{eq:eq6}
  \mathbf{H}(\mathbf{Y} |  \mathbf{X} = \bm{X}^-) = - \sum_{y \in \mathcal{Y}} \hat{p}(\mathbf{Y} | \mathbf{X}^-) \log \hat{p}(\mathbf{Y} | \mathbf{X}^-) \; .
\end{equation}
Over irrelevant regions, the model is required to be \emph{unable to decide} which image class to predict since there is no evidence to support any class. This can be seen as a high uncertainty in the model decision. Therefore, we consider maximizing the entropy of Eq.\ref{eq:eq6}. The later reaches its maximum at the uniform distribution \citep{coverentropy2006}. Thus, the inability of the model to decide is reached since each class is \emph{equiprobable}. An alternative to maximizing Eq.\ref{eq:eq6} is to use a supervised target distribution since it is already known (i.e., uniform distribution). To this end, we consider ${q}$ as a uniform distribution,
${
  q(\mathbf{Y} = y |  \mathbf{X} = \bm{X}^-_i) = 1/c \; , \forall y \in \mathcal{Y} \; ,
}
$
and caste a supervised learning setup using a cross-entropy between $q$ and ${\hat{p}}$ over ${\bm{X}^-}$,
\begin{align}
 \mathbf{H}(q_i, \hat{p}_i)^- &= - \sum_{y \in \mathcal{Y}}  q(\mathbf{Y} = y |  \mathbf{X} =  \bm{X}^-_i) \; \log \hat{p}(\mathbf{Y} = y |  \mathbf{X} = \bm{X}^-_i)
  = - \frac{1}{c} \sum_{y \in \mathcal{Y}} \log \hat{p}(y |  \bm{X}^-_i) \label{eq:eq9} \; .
\end{align}
The minimum of Eq.\ref{eq:eq9} is reached when ${\hat{p}(\mathbf{Y} |  \mathbf{X} = \bm{X}^-_i)}$ is uniform, thus, Eq.\ref{eq:eq6} reaches its maximum. Now, we can write the total training loss to be minimized as,
\begin{equation}
\label{eq:eq9-1}
  \min \E_{(\bm{X}_i, y_i) \in\mathbb{D}}\big[\mathbf{H}(p_i, \hat{p}_i)^+ + \mathbf{H}(q_i, \hat{p}_i)^-\big] \; .
\end{equation}
The posterior probability ${\hat{p}}$ is modeled using a classifier ${\mathcal{C}(. \;, \bm{\theta}_{\mathcal{C}})}$ with a set of parameters ${\bm{\theta}_{\mathcal{C}}}$; it can operate either on ${\bm{X}^+_i}$ or ${\bm{X}^-_i}$. The binary mask ${\bm{R}^+_i}$ (and ${\bm{R}^-_i}$) is learned using another model ${\mathcal{M}(\bm{X}_i ;\; \bm{\theta}_{\mathcal{M}})}$ with a set of parameters ${\bm{\theta}_{\mathcal{M}}}$. In this work, both models are based on neural networks (fully convolutional networks \citep{LongSDcvpr15} in particular). The networks ${\mathcal{M}}$ and ${\mathcal{C}}$ can be seen as two parts of one single network ${\mathcal{G}}$ that localizes regions of interest using a binary mask, then classifies their content. Fig.\ref{fig:fig-4} illustrates the entire model.
Due to the depth of ${\mathcal{G}}$, ${\mathcal{M}}$ receives its supervised gradient based only on the error made by ${\mathcal{C}}$. In order to boost the supervised gradient at ${\mathcal{M}}$, and provide it with more hints to select the most discriminative regions with respect to the image class, we consider using a secondary classification task at the output of ${\mathcal{M}}$ to classify the input ${\bm{X}}$, following \citep{lee15apmlr}. ${\mathcal{M}}$ computes the posterior probability ${\hat{p}^s(\bm{Y} | \bm{X})}$ which is another estimate of ${p(\bm{Y} |\bm{X})}$. To this end, ${\mathcal{M}}$ is trained to minimize the cross-entropy between $p$ and ${\hat{p}^s}$,
\begin{equation}
  \label{eq:eq11}
  \mathbf{H}(p_i, \hat{p}_i^s) = - \log \hat{p}^s(\mathbf{Y}= y_i | \mathbf{X} = \bm{X}_i) \; .
\end{equation}
The total training loss to minimize is formulated as,
\begin{equation}
  \label{eq:eq11-1}
  \min_{\{\bm{\theta}_{\mathcal{M}}, \bm{\theta}_{\mathcal{C}}\}} \E_{(\bm{X}_i, y_i) \in\mathbb{D}}\big[\mathbf{H}(p_i, \hat{p}_i)^+ + \mathbf{H}(q_i, \hat{p}_i)^- + \mathbf{H}(p_i, \hat{p}_i^s)\big] \; .
\end{equation}

\textbf{Mask computation and recursive erasing:}
The mask ${\bm{R}^+}$ is computed using the last feature maps of ${\mathcal{M}}$ which contains high abstract descriminative activations. We note such feature maps by a tensor ${\bm{A}_i \in \R^{c\times h^{\prime}\times w^{\prime}}}$ that contains a spatial map for each class. ${\bm{R}^+_i}$ is computed by aggregating the spatial activation of all the classes as,
${
\bm{T}_i = \sum_{k=1}^c \hat{p}^s(\mathbf{Y}=k | \mathbf{X} = \bm{X}_i) \; \bm{A}_i(k) \; ,
}
$
where ${\bm{T}_i \in \R^{h^{\prime}\times w^{\prime}}}$ is the continuous downsampled version of ${\bm{R}^+_i}$, and ${\bm{A}_i(k)}$ is the feature map of the class ${k}$ of the input ${\bm{X}_i}$. At convergence, the posterior probability of the winning class is pushed toward $1$ while the rest is pushed down to ${0}$. This leaves only the feature map of the winning classe. ${\bm{T}_i}$ is upscaled using \emph{interpolation} (Sec.\ref{sub:interpolation-sup-mat}) to ${\bm{T}{\uparrow}_i \in \R^{h \times w}}$ which has the same size as the input ${\bm{X}}$, then pseudo-thresholded using a sigmoid function to obtain  a pseudo-binary ${\bm{R}^+_i}$,
\begin{equation}
  \label{eq:eq13}
  p_{\mathbf{M}}(\mathbf{m} = 1| \bm{X}_i, (r, z)) = 1/(1 + \exp(- \omega \times (\bm{T}{\uparrow}_i(r, z) - \sigma^{\prime}))) \; ,
\end{equation}
where ${\omega}$ is a constant scalar that ensures that the sigmoid approximately equals to $1$ when ${\bm{T}{\uparrow}_i(r, z)}$ is larger than ${\sigma^{\prime}}$, and approximately equals to $0$ otherwise. At this point, ${\bm{R}^-}$ may still contain discriminative regions. To alleviate this issue, we propose a learning incremental and recursive erasing approach that drives ${\mathcal{M}}$ to mine complete discriminative regions. The mining algorithm is consistent, sample dependent, it has a maximum recursion depth ${u}$, associates trust coefficients to each recursion, integrated within the backpropagation, operates only during training, and has a practical implementation. Due to space limitation, we left it in the supplementary material (Sec.\ref{sub:recursive-erasing-sup-mat}).

\section{Results and analysis}
\label{sec:experiments}
Our experiments focus simultaneously on classification and pointwise localization tasks. Thus, we consider datasets that provide both image and pixel-level labels for evaluation. Particularly, the following three datasets are considered: GlaS in medical domain, and CUB-200-2011 and Oxford flower 102 on natural scene images.
  \begin{inparaenum}[(1)]
  \item GlaS dataset, one of the rare medical datasets that fits our scenario \citep{rony2019weak-loc-histo-survey}, was provided in the 2015 Gland Segmentation in Colon Histology Images Challenge Contest\footnote{GlaS: \href{https://warwick.ac.uk/fac/sci/dcs/research/tia/glascontest}{warwick.ac.uk/fac/sci/dcs/research/tia/glascontest}.} \citep{sirinukunwattana2017gland}. The main task of the challenge is gland segmentation of microscopic images. However, image-level labels were provided as well. The dataset is composed of 165 images derived from 16 Hematoxylin and Eosin (H\&E) histology sections of two grades (classes): benign, and malignant. It is divided into 84 samples for training, and 80 samples for test. Images have a high variation in term of gland shape/size, and overall H\&E stain. In this dataset, the glandes are the regions of interest that the pathologists use to prognosis the image grading of being benign or malignant.
  \item CUB-200-2011 dataset\footnote{CUB-200-2011: \href{http://www.vision.caltech.edu/visipedia/CUB-200-2011.html}{www.vision.caltech.edu/visipedia/CUB-200-2011.html}} \citep{WahCUB2002011} is a dataset for bird species with ${11,788}$ samples and ${200}$ species. Preliminary experiments were conducted on small version of this datatset where we selected randomly 5 species and build a small dataset with $150$ samples for training, and $111$ for test; referred to in this work as CUB5. The entire dataset is referred to as CUB. In this dataset, the regions of interest are the birds.
  \item Oxford flower 102\footnote{Oxford flower 102: \href{http://www.robots.ox.ac.uk/~vgg/data/flowers/102/}{http://www.robots.ox.ac.uk/~vgg/data/flowers/102/}} \citep{nilsback2007delving} datatset is collection of 102 species (classes) of flowers commonly occurring in United Kingdom; referred to here as OxF. It contains a total of ${8,189}$ samples. We used the provided splits for training (${1,020}$ samples), validation (${1,020}$ samples) and test (${6,149}$ samples) sets. Regions of interest are the flowers which were segmented automatically. \end{inparaenum}
  In GlaS, CUB5 and CUB datasets, we randomly select $80\%$ of training samples for effective training, and $20\%$ for validation to perform early stopping. We provide in our public code the used splits and the deterministic code that generated them for the different datasets.

In all the experiments, image-level labels are used during training/evaluation, while pixel-level labels are used exclusively during evaluation. The evaluation is conducted at two levels: at image-level where the classification error is reported, and at the pixel-level where we report F1 score (Dice index) over the foreground (region of interest), referred to as F1$^+$. When dealing with binary data, F1 score is equivalent to Dice index. We report as well the F1 score over the background, referred to as F1$^-$, in order to measure how well the model is able to identify irrelevant regions. We compare our method to different methods of WSL. Such methods use similar pre-trained backbone (resent18 \citep{heZRS16}) for feature extraction and differ mainly in the final pooling layer: CAM-Avg uses average pooling \citep{zhou2016learning}, CAM-Max uses max-pooling \citep{oquab2015object}, CAM-LSE uses an approximation to maximum \citep{PinheiroC15cvpr,sun2016pronet}, Wildcat uses the pooling in \citep{durand2017wildcat}, Grad-CAM \citep{selvaraju2017grad}, and Deep MIL is the work of \citep{ilse2018attention} with adaptation to multi-class. We use supervised segmentation using U-Net \citep{Ronneberger-unet-2015} as an upper bound of the performance for pixel-level evaluation (Full sup.). As a simple baseline, we use a mask full of 1 with the same size of the image as a constant prediction of regions of interest to show that F1$^+$ alone is not an efficient metric to evaluate pixel-level localization particularly over GlaS set (All-ones, Tab.\ref{tab:tab3-pl}). In our method, ${\mathcal{M}}$ and ${\mathcal{C}}$ \emph{share} the same pre-trained backbone (resnet101 \citep{heZRS16}) to avoid \emph{overfitting} while using \citep{durand2017wildcat} as a pooling function. All methods are trained using stochastic gradient descent using momentum. In our approach, we use the same hyper-parameters over all datasets, while other methods require adaptation to each dataset. We provide the datasets splits, more experimental details, and visual results in the supplementary material (Sec.\ref{sec:experiments-sup-mat}). Our reproducible code is publicly available\footnote{\href{https://github.com/sbelharbi/wsol-min-max-entropy-interpretability}{https://github.com/sbelharbi/wsol-min-max-entropy-interpretability}}.

\begin{figure}[h!]
  \centering
  \includegraphics[width=1.\linewidth]{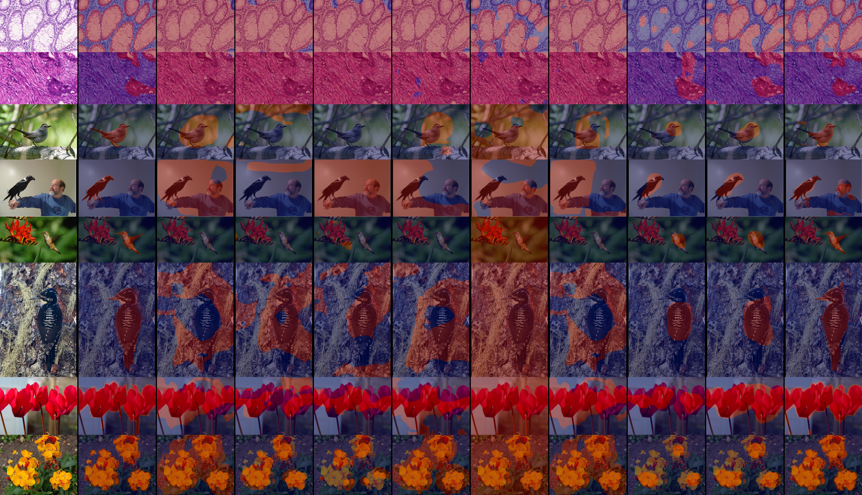} \\
  \includegraphics[width=1.\linewidth]{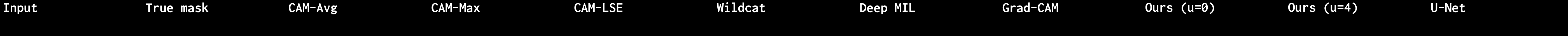}
  \caption{Visual comparison of the predicted binary mask of each method over GlaS,  CUB5, CUB, and OxF test sets. (Best visualized in color.) (See supplementary material for more samples.)}
  \label{fig:fig1-exp}
\end{figure}

A comparison of the obtained results of different methods, over all datasets, is presented in Tab.\ref{tab:tab3-il} and Tab.\ref{tab:tab3-pl} with visual results illustrated in Fig.\ref{fig:fig1-exp}. In Tab.\ref{tab:tab3-pl}, and compared to other WSL methods, our method obtains relatively similar F1$^+$ score; while it obtains large F1$^-$ over GlaS where it may be easy to obtain high F1$^+$ by predicting a mask full of 1 (Fig.\ref{fig:fig1-exp}). However, a model needs to be very selective in order to obtain high F1$^-$ score in order to localize tissues (irrelevant regions) where our model seems to excel at. Cub5 set seems to be more challenging due to the variable size (from small to big) of the birds, their view, the context/surrounding environment, and the few training samples. Our model outperforms all the WSL methods in both F1$^+$ and F1$^-$ with a large gap due mainly to its ability to discard non-discriminative regions which leaves it only with the region of interest, the bird in this case. While our model shows improvements in pointwise localization, it is still far behind full supervision.

Similar improvements are observed on CUB data. In the case of OxF dataset, our approach provides low F1$^+$ values compared to other WSL methods. However, the latter are not far from the performance of the All-ones that predicts a constant mask. Given the large size of flowers, predicting a mask that is active over all the image will easily lead to ${56.10\%}$ of F$^+$. The best WSL methods for OxF are only better than All-ones by $\sim2\%$, suggesting that such methods have predicted a full mask in many cases. In term of F1$^-$, our approach is better than all the WSL techniques. All methods achieve low classification error on GlaS which implies that it represents an easy classification problem. Surprisingly, the other methods seem to overfit on CUB5, while our model shows a robustness. The other methods outperform our approach on CUB and OxF, although ours is still in a competitive range to half WSL methods. Results obtained on both these datasets indicate that, compared to WSL methods, our approach is effective in terms of image classification and pointwise localization with more reliability in the latter.

Visual quality of our approach (Fig.\ref{fig:fig1-exp}) shows that the predicted regions of interest on GlaS agree with the doctor methodology of colon cancer diagnostics  where the glands are used as diagnostic tool. Additionally, it deals well with multi-instances when there are multiple glands within the image. On CUB5/CUB, our model succeeds to locate birds in order to predict its category which one may do in such task. We notice that the head, chest, tail, or body particular spots are often parts that are used by our model to decide a bird's species, which seems a reasonable strategy as well. On OxF dataset, we observe that our approach mainly locates the central part of pistil. When it is not enough, the model relies on the petals or on unique discriminative parts of the flower. In term of time complexity, the inference time of our model is the same as a standard fully convolutional network since the recursive algorithm is disabled during inference. However, one may expect a moderate increase in training time that depends mainly on the depth of the recursion (see Sec.\ref{subsub:exps-run-time-sup-mat}).

\begin{table}
  \caption{Image level performance over GlaS, CUB5, CUB, and OxF test sets.}
  \label{tab:tab3-il}
  \centering
  \small
  \resizebox{1.\linewidth}{!}{
  \begin{tabular}{l|l|l|l|l}
    \toprule
    & \multicolumn{4}{c}{Image level} \\
    Method & \multicolumn{4}{c}{Classification error (\%)} \\
    \cmidrule{2-5}
                      &  GlaS  &  CUB5 & CUB & OxF     \\
    \midrule
    CAM-Avg \citep{zhou2016learning}          & $\bm{0.00}$ &  $13.79$       & $24.62$       & $13.04$       \\
    CAM-Max  \citep{oquab2015object}          & $1.25$      &  $69.65$       & $30.30$       & $28.60$       \\
    CAM-LSE  \citep{PinheiroC15cvpr,sun2016pronet}     & $1.25$      &  $84.13$       & $28.44$       & $27.35$       \\
    Wildcat \citep{durand2017wildcat}           & $1.25$      &  $22.75$       & $\bm{22.12}$  & $13.01$       \\
    Deep MIL \citep{ilse2018attention}         & $2.50$      &  $12.41$       & $24.74$       & $\bm{12.14}$  \\
    Grad-CAM \citep{selvaraju2017grad}          & $\bm{0.00}$ &  $11.03$       & $24.62$       & $13.04$       \\
    \midrule
    Ours ($u=4$)         & $\bm{0.00}$  &  $\bm{10.34}$  & $26.73$       & $19.98$       \\
    \bottomrule
  \end{tabular}
  }
\end{table}

\begin{table}
  \caption{Pointwise localization performance over GlaS, CUB5, CUB, and OxF test sets.}
  \label{tab:tab3-pl}
  \centering
  \small
  \resizebox{1.\linewidth}{!}{
  \begin{tabular}{l|l|l|l|l|l|l|l|l}
    \toprule
    &  \multicolumn{8}{c}{Pixel level} \\
    Method &  \multicolumn{4}{c|}{F1$^+$ (\%)} &  \multicolumn{4}{c}{F1$^-$ (\%)}\\
    \cmidrule{2-9}
                      &  GlaS    &   CUB5 & CUB & OxF    &    GlaS   &  CUB5  & CUB & OxF \\
    \midrule
    All-ones          &   $66.01$   & $23.72$& $22.16$  & $56.10$    &  $00.00$  &  $00.00$ & $00.00$ & $00.00$ \\
    CAM-Avg \citep{zhou2016learning}          &   $66.90$   & $35.25$ & $32.24$  & $\bm{58.37}$  &  $17.88$  &  $68.44$& $82.87$ & $68.70$ \\
    CAM-Max  \citep{oquab2015object}             &  $66.00$   & $5.46$ & $35.41$  & $40.03$  &  $26.32$  &  $75.52$ & $91.62$ & $73.81$\\
    CAM-LSE  \citep{PinheiroC15cvpr,sun2016pronet}        &  $66.05$   & $8.00$ & $35.79$  & $40.65$  &  $27.93$  &  $77.21$ & $91.62$ & $73.07$\\
    Wildcat \citep{durand2017wildcat}        &  $67.21$   & $36.05$ & $37.91$  & $52.33$  &  $22.96$  &  $75.62$ & $89.09$ & $74.15$\\
    Deep MIL \citep{ilse2018attention}       &  $68.52$   & $29.70$  & $22.19$  & $56.10$ &  $41.34$  &  $37.59$ & $0.31$ & $0.0$\\
    Grad-CAM \citep{selvaraju2017grad}         &  $66.30$   & $36.91$  & $32.24$  & $\bm{58.37}$ &  $21.30$  &  $69.55$ & $82.87$ & $68.70$\\
    \midrule
    Ours ($u=4$)      &  $\bm{72.54}$   & $\bm{52.97}$ & $\bm{51.05}$  & $43.35$  & $\bm{66.51}$   &  $\bm{90.69}$ & $\bm{91.86}$ & $\bm{75.77}$\\
    \midrule
    \textbf{Full sup.: U-Net \citep{Ronneberger-unet-2015} }    &  $90.19$   & $60.06$ & $92.09$  & $88.81$  & $88.52$   &  $93.73$ & $98.97$ & $92.37$\\
    \bottomrule
  \end{tabular}
  }
\end{table}

\section{Conclusion}
\label{sec:conclusion}
In this work, we present a novel approach for WSL at pixel-level where we impose learning relevant and irrelevant regions within the model with the aim to reduce false positive localization. Evaluated on three datasets, and compared to state of the art WSL methods, our approach shows its effectiveness in accurately localizing regions of interest with low false positive while maintaining a competitive classification error. This makes our approach more reliable in term of interpetability. As future work, we consider extending our approach to handle multiple classes within the image. Different constraints can be applied over the predicted mask, such as texture properties, shape, or other region constraints. Predicting bounding boxes instead of heat maps is considered as well since they can be more suitable in some applications where pixel-level accuracy is not required. Our recursive erasing algorithm can be further improved by using a memory-like mechanism that provides spatial information to prevent forgetting the previously spotted regions and promote localizing the entire region (Sec.\ref{sub:results-exp-sup-mat}).


\subsubsection*{Acknowledgments}
  This work was partially supported by the Natural Sciences and Engineering Research Council of Canada and the Canadian Institutes of Health Research.

\bibliography{bibliography}
\bibliographystyle{iclr2020_conference}

\newpage
\clearpage

\appendix
\section{The min-max entropy framework for WSL}
\label{sec:proposedmethod-sup-mat}

\subsection{Region completeness using incremental recursive erasing and trust coefficients}
\label{sub:recursive-erasing-sup-mat}
\begin{figure}[h!]
  \center
\includegraphics[width=1.\linewidth]{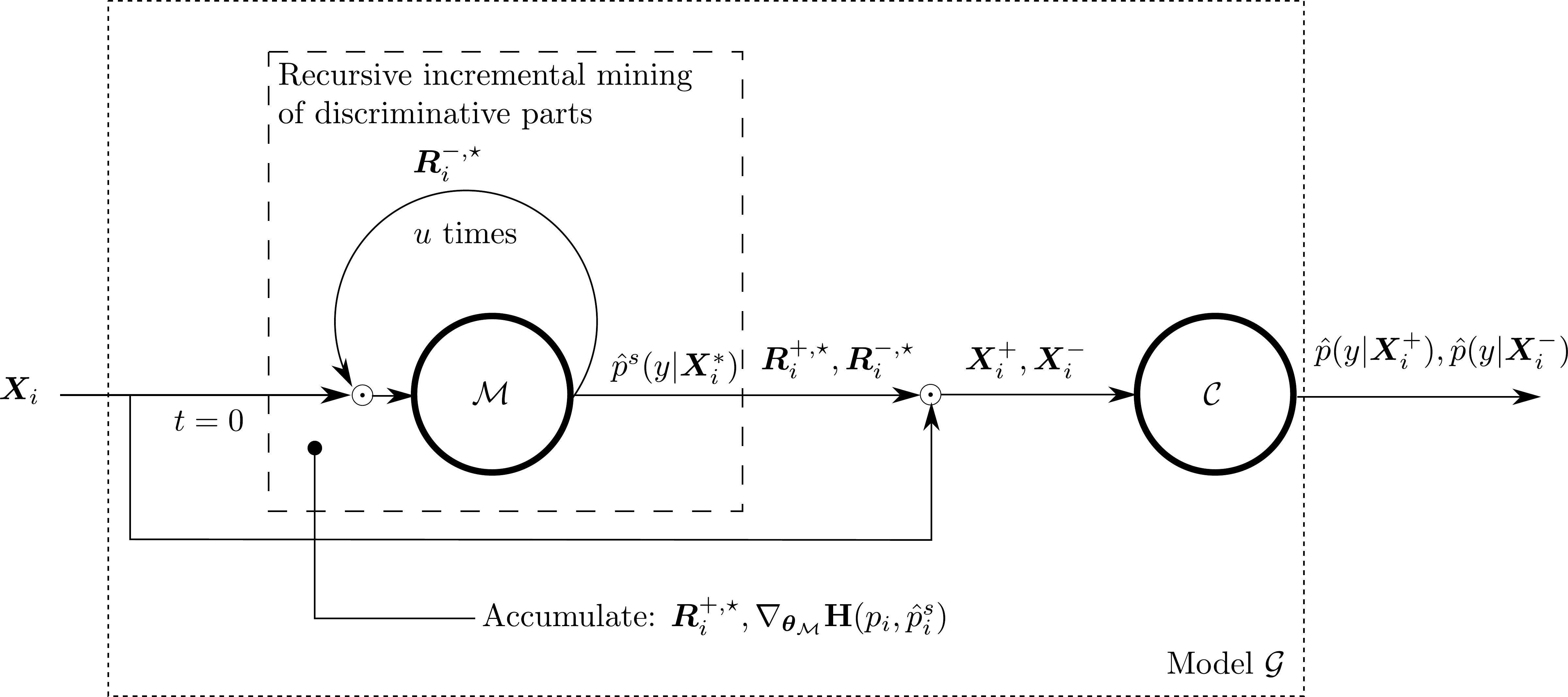}
\caption{Illustration of the implementation of the proposed recursive incremental mining of disciminative parts within the backpropagation. The recursive mining is performed only during training.}
\label{fig:fig-5-sup-mat}
\end{figure}

Deep classification models tend to rely on small discriminative regions \citep{kim2017two,SinghL17,zhou2016learning}. Thus, in our proposal, ${\bm{R}^-}$ may still contain discriminative parts. Following \citep{kim2017two,LiWPE018CVPR,pathak2015constrained,SinghL17}, and in particular \citep{wei2017object}, we propose a learning incremental and recursive erasing approach that drives  ${\mathcal{M}}$ to seek complete discriminative regions. However, in the opposite of \citep{wei2017object} where such mining is done offline, we propose to incorporate the erasing within the backpropagation using an efficient and practical implementation. This allows ${\mathcal{M}}$ to \emph{learn} to seek discriminative parts. Therefore, erasing during inference is unnecessary. Our approach consists in applying  ${\mathcal{M}}$ recursively  before applying ${\mathcal{C}}$ within the same forward. The aim of the recursion, with maximum depth $u$, is to mine more discriminative parts within the non-discriminative regions of the image masked by ${\bm{R}^-}$. We accumulate all discriminative parts in a temporal mask ${\bm{R}^{+,\star}}$. At each recursion, we mine the most discriminative part, that has been correctly classified by ${\mathcal{M}}$, and accumulate it in ${\bm{R}^{+,\star}}$. However, with the increase of $u$, the image may run out of discriminative parts. Thus, ${\mathcal{M}}$ is forced, unintentionally, to consider non-discriminative parts as discriminative. To alleviate this risk, we introduce \emph{trust coefficients} that control how much we trust a mined discriminative region at each step $t$ of the recursion for each sample $i$ as follows,
\begin{equation}
  \label{eq:eq14}
  \bm{R}^{+,\star}_i \coloneqq \max(\bm{R}^{+,\star}_i, \Psi(t, i) \; \bm{R}^{+, t}_i) \; ,
\end{equation}
where ${\Psi(t, i) \in \R^+}$ computes the trust of the current mask of the sample $i$ at the step $t$ as follows,
\begin{equation}
  \label{eq:eq15}
  \forall t \geq 0, \quad \Psi(t, i) = \exp^{\frac{-t}{\sigma}} \; \Gamma(t, i) \; ,
\end{equation}
where ${\exp^{\frac{-t}{\sigma}}}$ encodes the overall trust with respect to the current step of the recursion. Such trust is expected to decrease with the depth of the recursion \citep{bel16}.
${\sigma}$ controls the slop of the trust function. The second part of Eq.\ref{eq:eq15} is computed with respect to each sample. It quantifies how much we trust the estimated mask for the current sample $i$,
\begin{equation}
  \label{eq:eq16}
    \Gamma(t, i) =
    \begin{cases*}
      \hat{p}^s(\mathbf{Y}=y_i | \mathbf{X} = \bm{X}_i \odot \bm{R}^{-,\star}_i)  & if $\hat{y}_i = y_i $ and $\mathbf{H}(p_i, \hat{p}^s_i)_t \leq \mathbf{H}(p_i, \hat{p}^s_i)_0$ \; ,  \\
      0        & otherwise \; .
    \end{cases*}
  \end{equation}
In Eq.\ref{eq:eq16}, ${\mathbf{H}(p_i, \hat{p}^s_i)_t}$ is computed over ${(\bm{X}_i \odot \bm{R}^{-,\star}_i)}$. Eq.\ref{eq:eq16} ensures that at a step $t$, for a sample $i$, the current mask is trusted only if  ${\mathcal{M}}$ correctly classifies the erased image, and does not increase the loss. The first condition ensures that the accumulated discriminative regions belong to the same class, and more importantly, the true class. Moreover, it ensures that ${\mathcal{M}}$ does not change its class prediction through the erasing process. This introduces a \emph{consistency} between the mined regions across the steps and avoids mixing discriminative regions of different classes. The second condition ensures maintaining, at least, the same confidence in the predicted class compared to the first forward without erasing (${t=0}$).
The given trust in this case is equal to the probability of the true class. The regions accumulator is initialized to zero at ${t=0, \bm{R}^{+,\star}_i = \{0\}^{h\times w}}$ at each forward in ${\mathcal{G}}$. ${\bm{R}^{+,\star}_i}$ is not maintained through epoches; ${\mathcal{M}}$ starts over each time processing the sample $i$. This prevents accumulating incorrect regions that may occur at the beginning of the training. In order to automatize when to stop erasing, we consider a maximum depth of the recursion $u$. For a mini-batch, we keep erasing as along as we do not reach $u$ steps of erasing, and there is at least one sample with a trust coefficient non-zero (Eq.\ref{eq:eq16}). Once a sample is assigned a zero trust coefficient, it is maintained zero all along the erasing (Eq.\ref{eq:eq14})(Fig.\ref{fig:fig-5-sup-mat}). Direct implementation of Eq.\ref{eq:eq14} is not practical since performing a recursive computation on a large model ${\mathcal{M}}$ requires a large memory that increases with the depth $u$. To avoid such issue, we propose a practical implementation using gradient accumulation at ${\mathcal{M}}$ through the loss Eq.\ref{eq:eq11}; such implementation requires the same memory size as in the case without erasing. An illustration of our proposed recursive erasing algorithm is provided in Fig.\ref{fig:fig-5-sup-mat}. Alg.\ref{alg:alg0-sup-mat} illustrates our implementation using accumulated gradient through the backpropagation within the localizer ${\mathcal{M}}$. We note that this erasing algorithm is performed only during training.

\begin{algorithm}[h!]
    \caption{Practical implementation of our incremental recursive erasing approach during training for one epoch (or one mini-batch) using gradient accumulation.}
    \label{alg:alg0-sup-mat}
    \begin{algorithmic}[1]
      \State \textbf{Input:} $\mathcal{G}, \mathbb{D}, u, \sigma$.
      \State Initialization: ${\nabla_{\{\bm{\theta}_{\mathcal{M}}, \bm{\theta}_{\mathcal{C}}\}} \big[\mathbf{H}(p_i, \hat{p}_i)^+ + \mathbf{H}(q_i, \hat{p}_i)^- + \mathbf{H}(p_i, \hat{p}_i^s)\big] = \nabla_{\{\bm{\theta}_{\mathcal{M}}, \bm{\theta}_{\mathcal{C}}\}}^{\mathcal{G}} = \bm{0}}$.
      \For{${(\bm{X}_i, y_i) \in \mathbb{D}}$}
      \State Initialization: ${\bm{R}^{+,\star}_i = \bm{0}, \nabla_{\bm{\theta}_{\mathcal{M}}}\mathbf{H}(p_i, \hat{p}^s_i) = \bm{0}, t=0}$, stop = False.
      \State Make a copy of ${\bm{X}_i}$: ${\bm{X}^{\star}_i}$.
      \State \texttt{\# Perform the recursion. Accumulate gradients, and masks.}
      \While {$t \leq u$ and stop is False}
        \State Forward ${\bm{X}^{\star}_i}$ in ${\mathcal{M}}$.
        \State Compute ${\bm{R}^{+,t}_i, \bm{R}^{-,t}_i}, \hat{p}^s(y | \bm{X}^{\star}_i), \mathbf{H}(p_i, \hat{p}^s_i)_t, \Psi(t, i)$.
        \If{$\Psi(t, i) \neq 0$}
        \State Update accumulative mask ${\bm{R}^{+,\star}_i}$. (Eq.\ref{eq:eq14})
        \State Accumulate gradient: ${\nabla_{\bm{\theta}_{\mathcal{M}}}\mathbf{H}(p_i, \hat{p}^s_i) \pluseq  \nabla_{\bm{\theta}_{\mathcal{M}}}\mathbf{H}(p_i, \hat{p}^s_i)_t}$
        \State Erase the discriminative parts: ${\bm{X}^{\star}_i \coloneqq \bm{X}^{\star}_i \odot \bm{R}^{-,\star}_i}$.
        \Else
        \State stop = True.
        \EndIf
      \EndWhile
      \State Compute: ${\bm{X}^+_i = \bm{X}_i \odot \bm{R}^{+,\star}_i, \bm{X}^-_i = \bm{X}_i \odot \bm{R}^{-,\star}_i}$.
      \State Forward ${\bm{X}^+_i, \bm{X}^-_i}$ in ${\mathcal{C}}$.
      \State Compute: ${\mathbf{H}(p_i, \hat{p}_i)^+, \mathbf{H}(q_i, \hat{p}_i)^-, \nabla_{\bm{\theta}_{\mathcal{C}}} \Big[\mathbf{H}(p_i, \hat{p}_i)^+, \mathbf{H}(q_i, \hat{p}_i)^-\Big]}$.
      \State Update the total gradient: ${\nabla_{\{\bm{\theta}_{\mathcal{M}}, \bm{\theta}_{\mathcal{C}}\}}^{\mathcal{G}} \pluseq \nabla_{\bm{\theta}_{\mathcal{C}}} \Big[\mathbf{H}(p_i, \hat{p}_i)^+, \mathbf{H}(q_i, \hat{p}_i)^- \Big] +  \nabla_{\bm{\theta}_{\mathcal{M}}}\mathbf{H}(p_i, \hat{p}^s_i)}$.
      \EndFor
      \State Normalize total gradient: ${\nabla_{\{\bm{\theta}_{\mathcal{M}}, \bm{\theta}_{\mathcal{C}}\}}^{\mathcal{G}} \diveq n}$. Update ${\bm{\theta}_{\mathcal{M}}, \bm{\theta}_{\mathcal{C}}}$ using ${\nabla_{\{\bm{\theta}_{\mathcal{M}}, \bm{\theta}_{\mathcal{C}}\}}^{\mathcal{G}}}$.
      \State \textbf{Output:} ${\mathcal{G}}$ updated.
    \end{algorithmic}
\end{algorithm}

\subsection{Note on interpolation (Eq.\ref{eq:eq13})}
\label{sub:interpolation-sup-mat}
In most neural networks libraries (Pytorch (pytorch.org), Chainer (chainer.org)), the upsacling operations using interpolation/upsamling have a non-deterministic backward. This makes training unstable due to the non-deterministic gradient; and makes reproducibility  impossible as well. To avoid such issues, we detach the upsacling operation, in Eq.\ref{eq:eq13}, from the training graph and consider it as input data for ${\mathcal{C}}$.

\section{Results and analysis}
\label{sec:experiments-sup-mat}

In this section, we provide more details on our experiments, analysis, and discuss some of the drawbacks of our approach. We took many precautions to make the code reproducible for our model up to Pytorch's terms of reproducibility. Please see the \verb+README.md+ file for the concerned section in the code. We checked reproducibility up to a precision of $10^{-16}$. All our experiments were conducted using the seed $0$. We run all our experiments over one GPU with 12GB\footnote{Our code supports multiGPU, and Batchnorm synchronization with our own support to reproducibility.}, and an environment with 10 to 64 GB of RAM (depending on the size of the dataset). Finally, this section shows more visual results, analysis, training time, and drawbacks.

\subsection{Datasets}
\label{sub:datasets-exp-sup-mat}
\begin{figure}[h!]
    \centering
    \includegraphics[width=1.\linewidth]{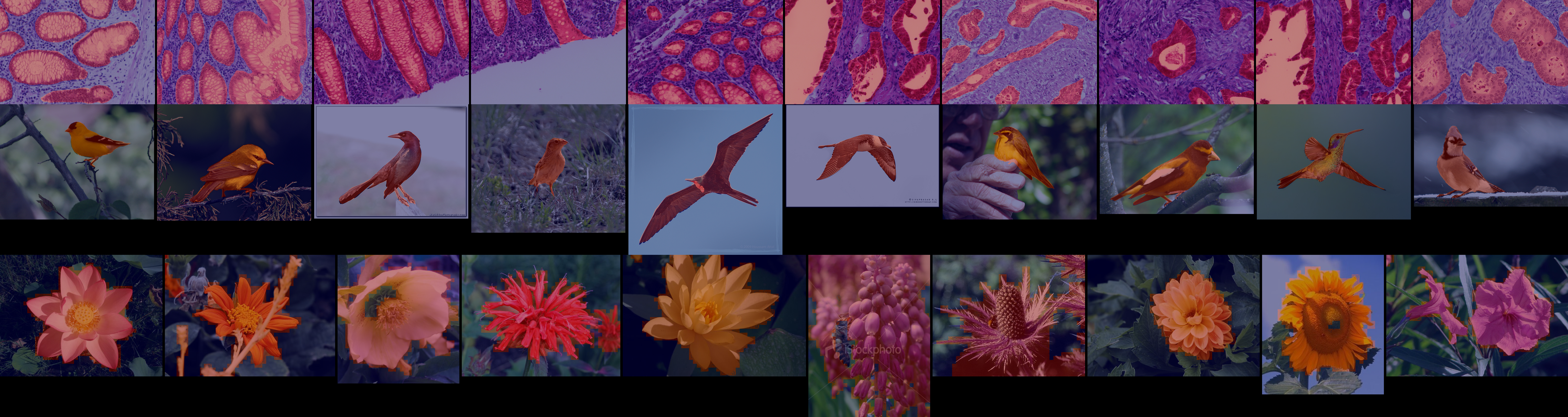}
    \caption{\textbf{Top row}: GlaS dataset: test set examples of different classes with the gland segmentation. The decidable regions are the glands while the undecidable regions are the leftover tissues. Glands have different shapes, size, context. They can be multi-instance. Images have variable H\&E stain.  \citep{sirinukunwattana2017gland}. \textbf{Middle row}: CUB dataset: test set examples of randomly selected classes. The decidable regions are the birds while the undecidable regions are the leftover surrounding environment. Birds have different sizes, position/view, appearance, context. \citep{WahCUB2002011}. \textbf{Bottom row}: Oxford flower 102 dataset: test samples of randomly selected classes. The decidable regions are the flowers while the undecidable ones are the surrounding environment. \citep{nilsback2007delving}  (Best visualized in color.)}
    \label{fig:fig-3-sup-mat}
\end{figure}

We provide in Fig.\ref{fig:fig-3-sup-mat} some samples from each dataset's test set along with their mask that indicates the region of interest.
As we mentioned in Sec.\ref{sec:experiments}, we consider a subset from the original CUB-200-2011 dataset for preliminary experiments, and we referred to it as CUB5. To build it, we select, randomly, 5 classes from the original dataset. Then, pick all the corresponding samples of each class in the provided train and test set to build our train and test set (CUB5). Then, we build the effective train set, and validation set by taking randomly $80\%$, and the left $20\%$ from the train set of CUB5, respectively. We provide the splits, and the code used to generate them. Our code generates the following classes:

\begin{minipage}{\textwidth}
  \begin{enumerate}
      \item \verb+019.Gray_Catbird+
      \item \verb+099.Ovenbird+
      \item \verb+108.White_necked_Raven+
      \item \verb+171.Myrtle_Warbler+
      \item \verb+178.Swainson_Warbler+
  \end{enumerate}
\end{minipage}

\subsection{Experiments setup}
\label{sub:exps-setup-exp-sup-mat}
The following is the configuration we used for our model over all the datasets:
\begin{description}
    \item[Data] \begin{inparaenum}
        \item Patch size (hxw): $480\times480$. (for training sample patches, however, for evaluation, use the entire input image).
        \item Augment patch using random rotation, horizontal/vertical flipping. (for CUB5 only horizontal flipping is performed).
        \item Channels are normalized using $0.5$ mean and $0.5$ standard deviation.
        \item For GlaS: patches are jittered using brightness=$0.5$, contrast=$0.5$, saturation=$0.5$, hue=$0.05$.
    \end{inparaenum}
    \item[Model] Pretrained resnet101 \citep{heZRS16} as a backbone with \citep{durand2017wildcat} as a pooling score with our adaptation, using $5$ modalities per class. We consider using dropout \citep{srivastava14a} (with value $0.75$ over GlaS and $0.85$ over CUB5, CUB, OxF over the final map of the pooling function right before computing the score). High dropout is motivated by \citep{GhiasiNIPS2018,SinghL17}. This allows to drop most discriminative parts at features with most abstract representation. The dropout is not performed over the final mask, but only on the internal mask of the pooling function. As for the parameters of \citep{durand2017wildcat}, we consider their $\alpha=0$ since most negative evidence is dropped, and use $kmax=kmin=0.09$. $u=0, u=4, \sigma=10, \sigma^{\prime}=0.5, \omega=8$. For evaluation, our predicted mask is binarized using a $0.5$ threshold to obtain exactly a binary mask. All our presented masks in this work follows this thresholding. Our F1$^+$, and F1$^-$ are computed over this binary mask.
    \item[Optimization]
    \begin{inparaenum}
        \item Stochastic gradient descent, with momentum $0.9$, with Nesterov.
        \item Weight decay of $1e-5$ over the weights.
        \item Learning rate of $0.001$ decayed by $0.1$ each $40$ epochs with minimum value of $1e-7$.
        \item Maximum epochs of 400.
        \item Batch size of $8$.
        \item Early stopping over validation set using classification error as a stopping criterion.
    \end{inparaenum}
\end{description}

Other WSL methods use the following setup with respect to each dataset:

\textbf{GlaS}:

\begin{description}
    \item[Data] \begin{inparaenum}
        \item Patch size (hxw): $416\times416$.
        \item Augment patch using random horizontal flip.
        \item Random rotation of one of: $0, 90, 180, 270$ (degrees).
        \item Patches are jittered using brightness=$0.5$, contrast=$0.5$, saturation=$0.5$, hue=$0.05$.
    \end{inparaenum}
    \item[Model]
    \begin{inparaenum}
        \item Pretrained resnet18 \citep{heZRS16} as a backbone.
    \end{inparaenum}
    \item[Optimization]
    \begin{inparaenum}
        \item Stochastic gradient descent, with momentum $0.9$, with Nesterov.
        \item Weight decay of $1e-4$ over the weights.
        \item 160 epochs
        \item Learning rate of $0.01$ for the first $80$, and of $0.001$ for the last $80$ epochs.
        \item Batch size of $32$.
        \item Early stopping over validation set using classification error/loss as a stopping criterion.
    \end{inparaenum}
\end{description}

\textbf{CUB5}:

\begin{description}
    \item[Data] \begin{inparaenum}
        \item Patch size (hxw): $448\times448$. (resized while maintaining the ratio).
        \item Augment patch using random horizontal flip.
        \item Random rotation of one of: $0, 90, 180, 270$ (degrees).
        \item Random affine transformation with degrees $10$, shear $10$, scale $(0.3, 1.5)$.
    \end{inparaenum}
    \item[Model] Pretrained resnet18 \citep{heZRS16} as a backbone.
    \item[Optimization]
    \begin{inparaenum}
        \item Stochastic gradient descent, with momentum $0.9$, with Nesterov.
        \item Weight decay of $1e-4$ over the weights.
        \item $90$ epochs.
        \item Learning rate of $0.01$ decayed every $30$ with $0.1$.
        \item Batch size of $8$.
        \item Early stopping over validation set using classification error/loss as a stopping criterion.
    \end{inparaenum}
\end{description}

\textbf{CUB/OxF}:

\begin{description}
    \item[Data] \begin{inparaenum}
        \item Patch size (hxw): $448\times448$. (resized while maintaining the ratio).
        \item Augment patch using random horizontal flip.
        \item Random rotation of one of: $0, 90, 180, 270$ (degrees).
        \item Random affine transformation with degrees $10$, shear $10$, scale $(0.3, 1.5)$.
    \end{inparaenum}
    \item[Model] Pretrained resnet18 \citep{heZRS16} as a backbone.
    \item[Optimization]
    \begin{inparaenum}
        \item Stochastic gradient descent, with momentum $0.9$, with Nesterov.
        \item Weight decay of $1e-4$ over the weights.
        \item $90$ epochs.
        \item Learning rate of $0.01$ decayed every $30$ with $0.1$.
        \item Batch size of $64$.
        \item Early stopping over validation set using classification error/loss as a stopping criterion.
    \end{inparaenum}
\end{description}

\subsection{Results}
\label{sub:results-exp-sup-mat}
In this section, we provide more visual results over the test set of each dataset.

Over GlaS dataset (Fig.\ref{fig:fig1-exp-sup-mat-1}, \ref{fig:fig1-exp-sup-mat-2}), the visual results show clearly how our model, with and without erasing, can handle multi-instance. Adding the erasing feature allows recovering more discriminative regions. The results over CUB5 (Fig.\ref{fig:fig3-exp-sup-mat-4}, \ref{fig:fig3-exp-sup-mat-5}, \ref{fig:fig3-exp-sup-mat-6}, \ref{fig:fig3-exp-sup-mat-7}, \ref{fig:fig3-exp-sup-mat-8}) while are interesting, they show a fundamental limitation to the concept of erasing in the case of one-instance. In the case of multi-instance, when the model spots one instance, then, erases it, it is more likely that the model will seek another instance which is the expected behavior. However, in the case of one instance, and where the discriminative parts are small, the first forward allows mainly to spot such small part and erase it. Then, the leftover may not be sufficient to discriminate. For instance, in CUB5, in many cases, the model spots only the head. Once it is hidden, the model is unable to  find other discriminative parts. A clear illustration to this issue is in Fig.\ref{fig:fig3-exp-sup-mat-4}, row 13. The model spots correctly the head, but was unable to spot the body while the body has similar texture, and it is located right near to the found head. We believe that the main cause of this issue is that the erasing concept \emph{forgets} where discriminative parts are located since the mining iterations are done independently from each other in a sens that the next mining iteration is unaware of what was already mined. Erasing algorithms seem to be missing this feature that can be helpful to localize the entire region of interest by seeking \emph{around} all the previously mined disciminative regions. In our erasing algorithm, once a region is erased, the model forgets about its location. Adding a memory-like, or constraints over the spatial distribution of the mined discriminative regions may potentially alleviate this issue. Another parallel issue of erasing algorithms is that once the most discriminative regions are erased it may not be possible to discriminate using the leftover regions. This may explain why our model was unable to spot other parts of the bird once its head is erased. Probably using soft-erasing (blur the pixel for example) can be more helpful than hard-erasing (set pixel to zero).

It is interesting to notice the strategy used by our model to localize some types of birds. In the case of the \verb+099.Ovenbird+, it relies on the texture of the chest (white doted with black), while it localizes the white spot on the bird neck in the case of \verb+108.White_necked_Raven+. One can notice as well that our model seems to be robust to small/occluded regions. In many cases, it was able to spot small birds in a difficult context where the bird is not salient.

Visual results over CUB and OxF are presented in Fig.\ref{fig:fig3-exp-sup-mat-9}, and Fig.\ref{fig:fig3-exp-sup-mat-10}, respectively.

\subsubsection{Impact of our recursive erasing algorithm on the  performance}
\label{subsun:entirecub}
\begin{figure}[h!]
  \includegraphics[width=1.\linewidth]{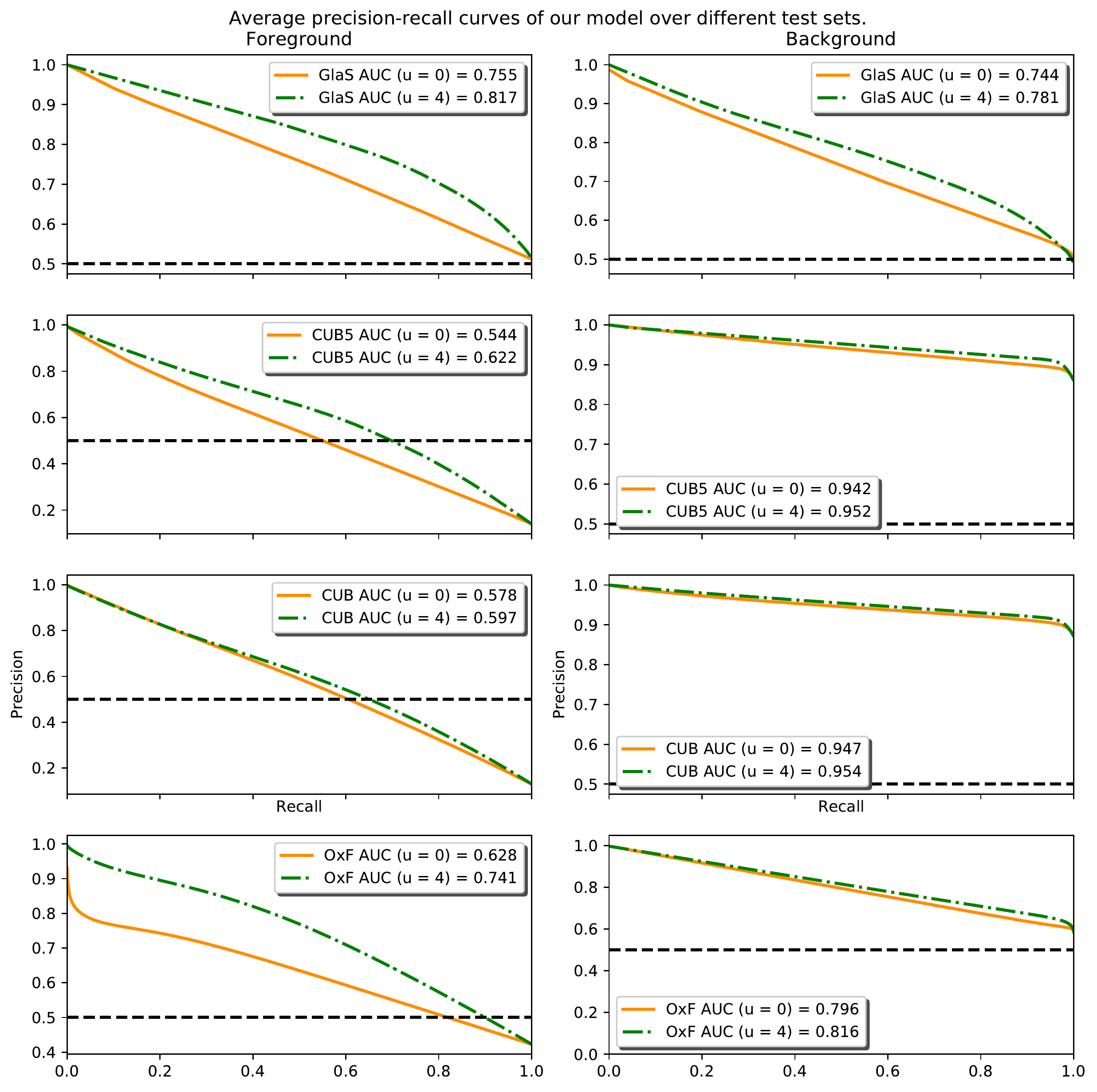}
  \cprotect\caption{\textbf{Average} precision-recall curve of the foreground and the background of our proposal using ${u=0, u=4}$ over each test set. To be able to compute an average curve, the recall axis is unified for all the images to the axis ${[0, 1]}$ with a step ${1e-3}$. Then, the precision axis is interpolated with respect to the recall axis.}
  \label{fig:fig-p-r-exp-sup-mat-1}
\end{figure}

\begin{figure}[h!]
  \includegraphics[width=1.\linewidth]{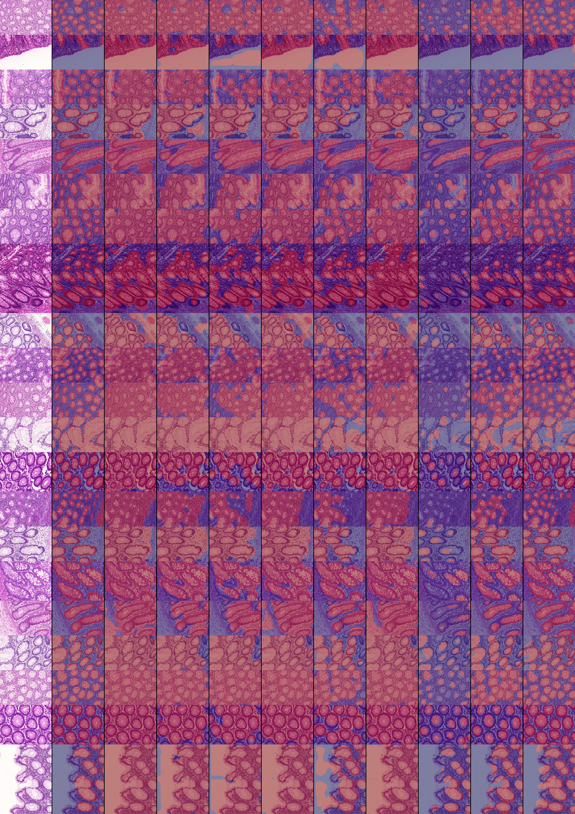} \\
  \includegraphics[width=1.\linewidth]{code}
  \cprotect\caption{Visual comparison of the predicted binary mask of each method over GlaS test set. Class: \verb+benign+ (Best visualized in color.)}
  \label{fig:fig1-exp-sup-mat-1}
\end{figure}

\begin{figure}[h!]
  \includegraphics[width=1.\linewidth]{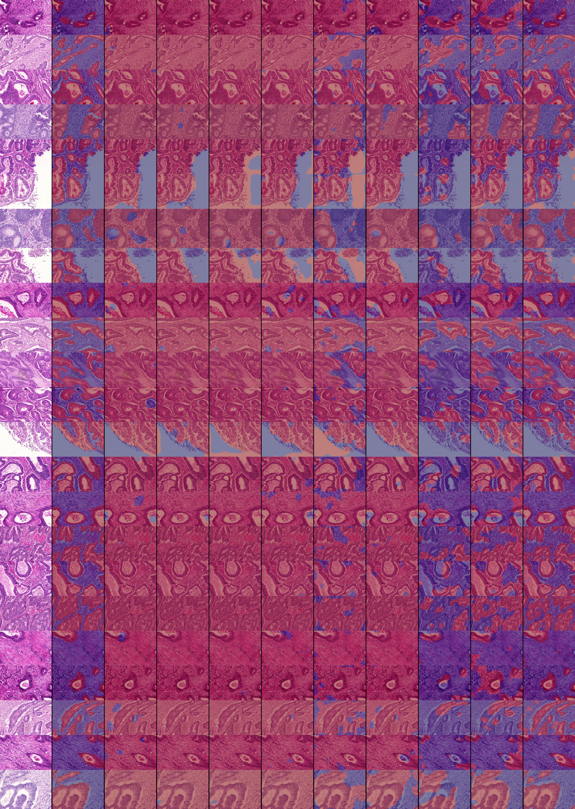} \\
  \includegraphics[width=1.\linewidth]{code}
  \cprotect\caption{Visual comparison of the predicted binary mask of each method over GlaS test set. Class: \verb+malignant+ (Best visualized in color.)}
  \label{fig:fig1-exp-sup-mat-2}
\end{figure}

\begin{figure}[h!]
\includegraphics[width=1.\linewidth]{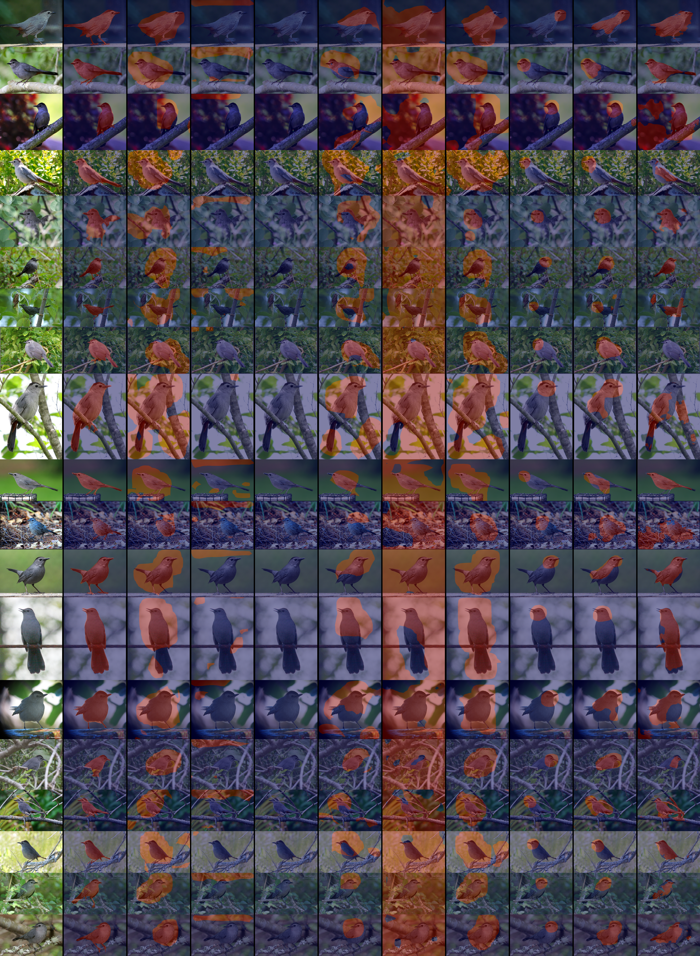} \\
\includegraphics[width=1.\linewidth]{code}
  \cprotect\caption{Visual comparison of the predicted binary mask of each method over CUB-200-2011 (CUB5) test sets. Species: \verb+019.Gray_Catbird+ (Best visualized in color.)}
  \label{fig:fig3-exp-sup-mat-4}
\end{figure}

\begin{figure}[h!]
\includegraphics[width=1.\linewidth]{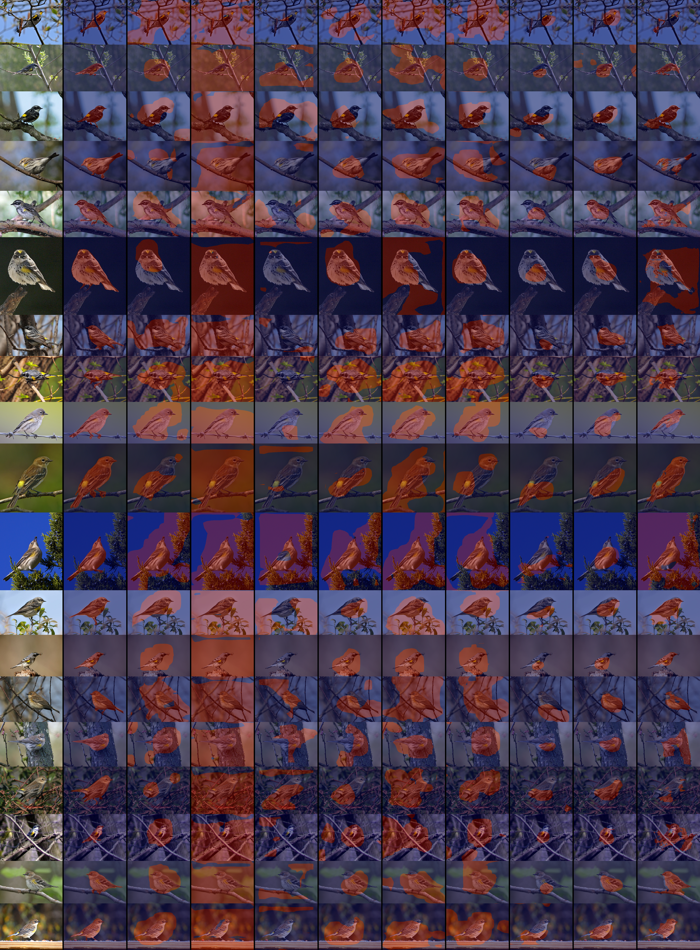} \\
\includegraphics[width=1.\linewidth]{code}
  \cprotect\caption{Visual comparison of the predicted binary mask of each method over CUB-200-2011 (CUB5) test sets. Species: \verb+171.Myrtle_Warbler+ (Best visualized in color.)}
  \label{fig:fig3-exp-sup-mat-5}
\end{figure}

\begin{figure}[h!]
\includegraphics[width=1.\linewidth]{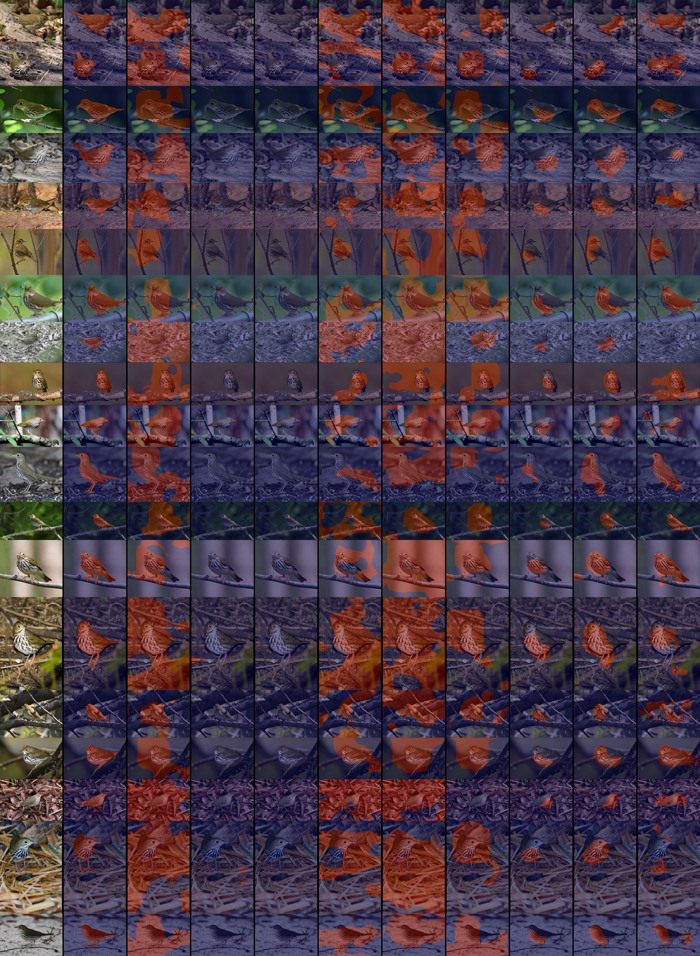} \\
\includegraphics[width=1.\linewidth]{code}
  \cprotect\caption{Visual comparison of the predicted binary mask of each method over CUB-200-2011 (CUB5) test sets. Species: \verb+099.Ovenbird+ (Best visualized in color.)}
  \label{fig:fig3-exp-sup-mat-6}
\end{figure}

\begin{figure}[h!]
\includegraphics[width=1.\linewidth]{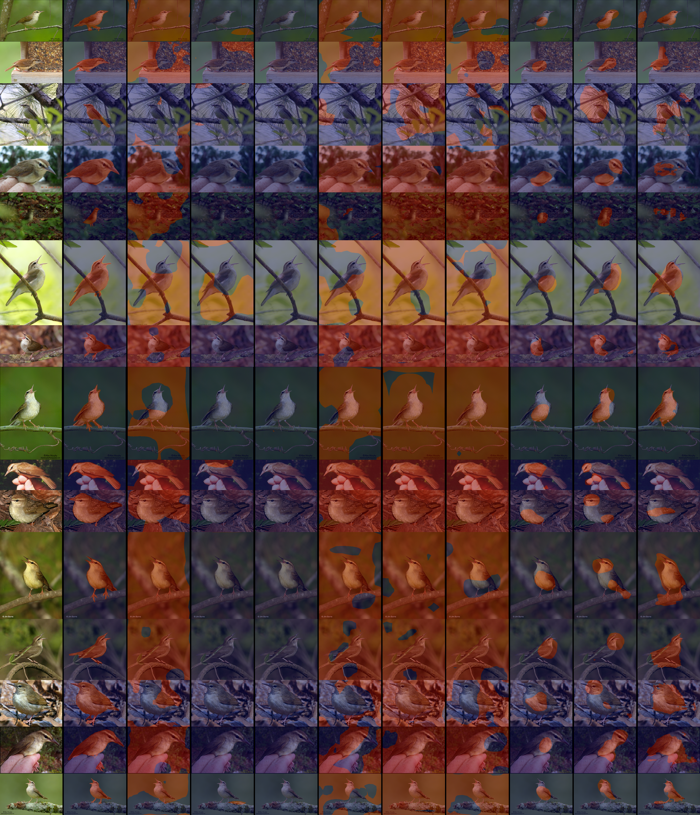} \\
\includegraphics[width=1.\linewidth]{code}
  \cprotect\caption{Visual comparison of the predicted binary mask of each method over CUB-200-2011 (CUB5) test sets. Species: \verb+178.Swainson_Warbler+ (Best visualized in color.)}
  \label{fig:fig3-exp-sup-mat-7}
\end{figure}

\begin{figure}[h!]
\includegraphics[width=1.\linewidth]{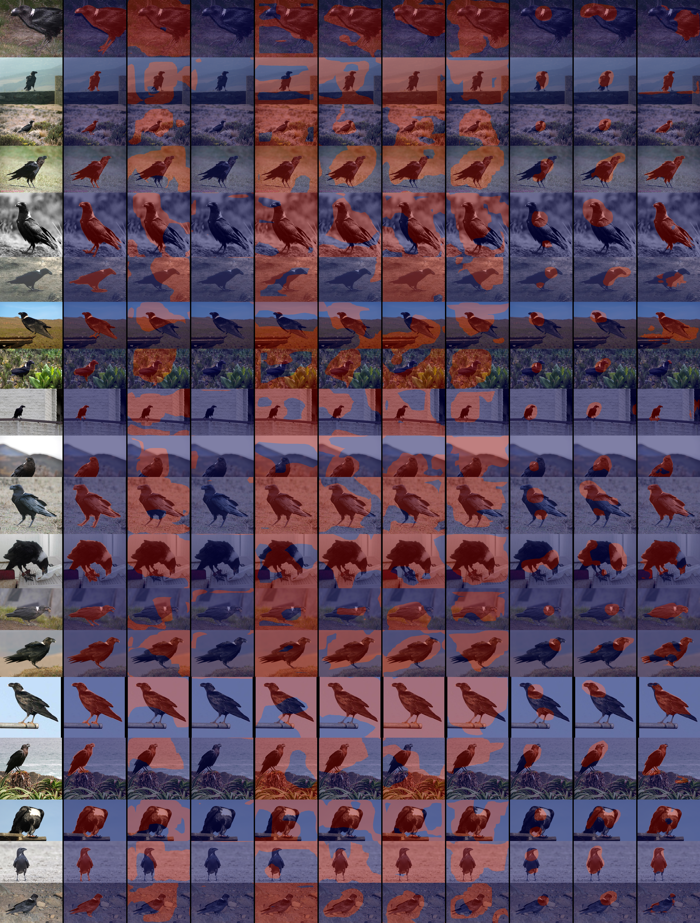} \\
\includegraphics[width=1.\linewidth]{code}
  \cprotect\caption{Visual comparison of the predicted binary mask of each method over CUB-200-2011 (CUB5) test sets. Species: \verb+108.White_necked_Raven+ (Best visualized in color.)}
  \label{fig:fig3-exp-sup-mat-8}
\end{figure}

\begin{figure}[h!]
\includegraphics[width=1.\linewidth]{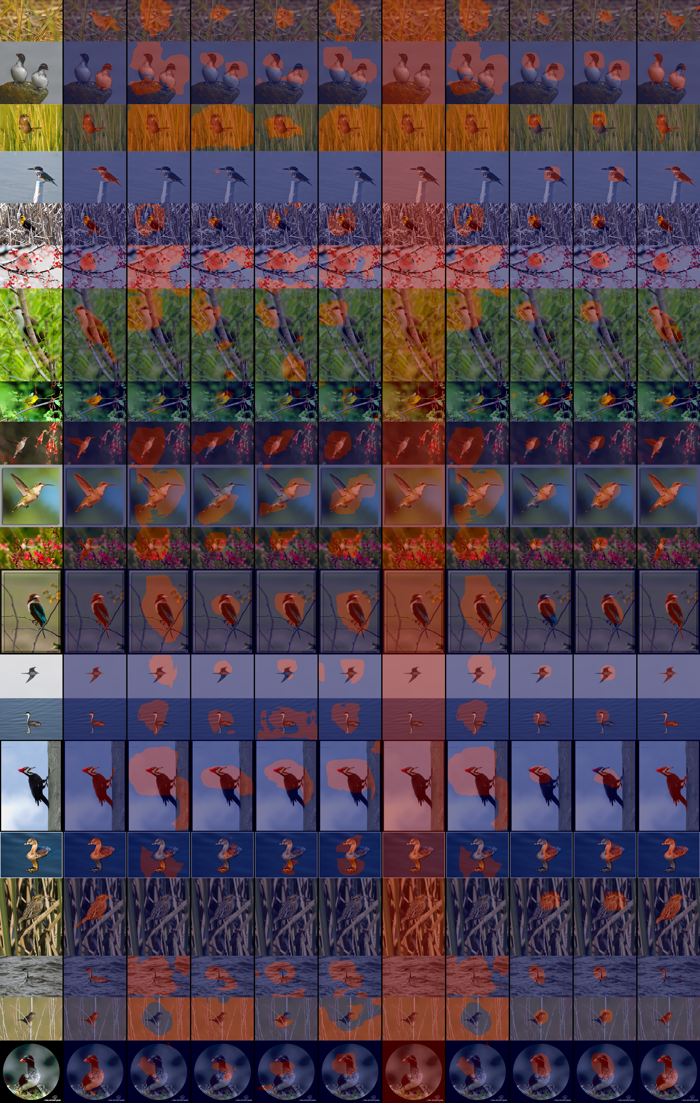} \\
\includegraphics[width=1.\linewidth]{code}
  \cprotect\caption{Visual comparison of the predicted binary mask of each method over CUB test sets. (Best visualized in color.)}
  \label{fig:fig3-exp-sup-mat-9}
\end{figure}

\begin{figure}[h!]
\includegraphics[width=1.\linewidth]{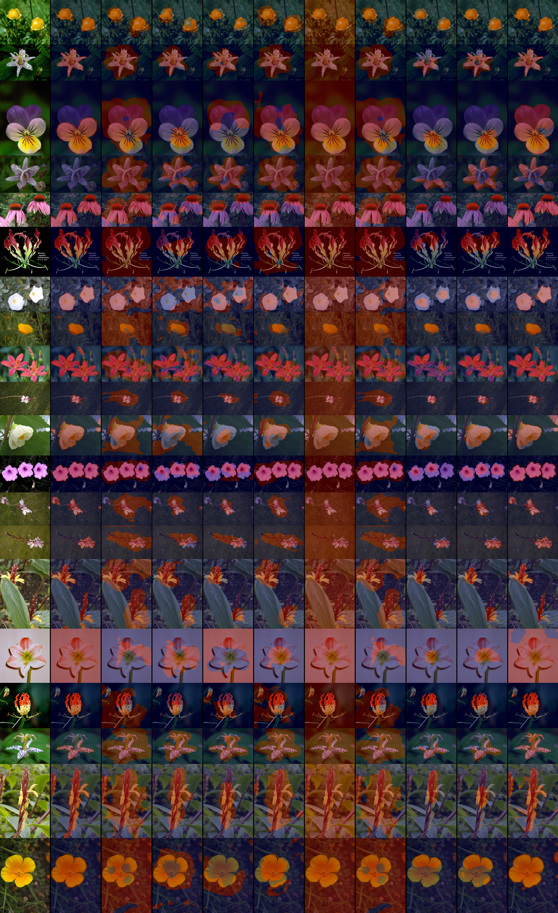} \\
\includegraphics[width=1.\linewidth]{code}
  \cprotect\caption{Visual comparison of the predicted binary mask of each method over oxF test sets. (Best visualized in color.)}
  \label{fig:fig3-exp-sup-mat-10}
\end{figure}

Tab.\ref{tab:tab4-1-sup-mat} and Tab.\ref{tab:tab4-2-sup-mat} show the boosting impact of our erasing recursive algorithm in both classification and pointwise localization performance. From Tab.\ref{tab:tab4-2-sup-mat}, we can observe that using our recursive algorithm adds a large improvement in F1$^+$ without degrading F1$^-$. This means that the recursion allows the model to correctly localize larger portions of the region of interest \emph{without} including false positive parts. The observed improvement in localization allows better classification error as observed in Tab.\ref{tab:tab4-1-sup-mat}. The localization improvement can be seen as well in the precision-recall curves in Fig.\ref{fig:fig-p-r-exp-sup-mat-1}.

\begin{table}[h!]
  \caption{Impact of our incremental recursive erasing algorithm over the classification error of our approach over GlaS, CUB5, CUB, and OxF test sets.}
  \label{tab:tab4-1-sup-mat}
  \centering
  \begin{tabular}{l|l|l|l|l}
    \toprule
    & \multicolumn{4}{c}{Image level} \\
    Ours & \multicolumn{4}{c}{Classification error (\%)} \\
    \cmidrule{2-5}
                      &  GlaS      &  CUB5           & CUB             &  OxF       \\
    \midrule
    $u=0$             & $1.25$     &  $19.31$        &  $\bm{26.54}$        & $25.15$  \\
    $u=4$            & $\bm{0.00}$ &  $\bm{10.34}$   &  $26.73$   & $\bm{19.98}$  \\
    \bottomrule
  \end{tabular}
\end{table}

\begin{table}[h!]
  \caption{Impact of our incremental recursive erasing algorithm over the pointwise localization performance of our approach over GlaS, CUB5, CUB, and OxF test sets.}
  \label{tab:tab4-2-sup-mat}
  \centering
  \begin{tabular}{l|l|l|l|l|l|l|l|l}
    \toprule
    & \multicolumn{8}{c}{Pixel level} \\
    Ours & \multicolumn{4}{c|}{F1$^+$ (\%)} &  \multicolumn{4}{c}{F1$^-$ (\%)}\\
    \cmidrule{2-9}
                      &      GlaS    &   CUB5    & CUB    &  OxF      &    GlaS    &  CUB5 & CUB &  OxF   \\
    \midrule
    $u=0$              & $39.99$    &  $40.07$  & $46.39$ &  $23.43$   &  $65.30$   & $89.99$   & $85.94$   &  $73.65$ \\
    $u=4$             & $\bm{72.54}$   &  $\bm{52.97}$ & $\bm{51.05}$ & $\bm{43.35}$ & $\bm{66.51}$ & $\bm{90.69}$ & $\bm{91.86}$ & $\bm{75.77}$ \\
    \bottomrule
  \end{tabular}
\end{table}

\subsubsection{Running time of our recursive erasing algorithm}
\label{subsub:exps-run-time-sup-mat}
Adding recursive computation in the backpropagation loop is expected to add an extra computation time. Tab.\ref{tab:tab4-sup-mat} shows the training time (of 1 run) of our model with and without recursion over identical computation resource. The observed extra computation time is mainly due to gradient accumulation (line 12. Alg.\ref{alg:alg0-sup-mat}) which takes the same amount of time as parameters' update (which is expensive to compute). The forward and the backward are practically fast, and take less time compared to gradient update. We do not compare the running between the datasets since they have different number/size of samples, and different pre-processing that it is included in the reported time. Moreover, the size of samples has an impact over the total time during the training over the validation set.

\begin{table}[h!]
  \caption{Comparison of training time, of 1 run, over 400 epochs over GlaS and CUB5 of our model using identical computation resources (NVIDIA Tesla V100 with 12GB memory) when using our erasing algorithm ($u=4$) and without using it ($u=0$).}
  \label{tab:tab4-sup-mat}
  \centering
  \begin{tabular}{lll}
    \toprule
    Model                    & GlaS        & CUB5 \\
    \midrule
    Ours ($u=0$)         &  49min    &  65min\\
    Ours ($u=4$)     &  90min ($\sim \times 1.83$)   &  141min ($\sim \times 2.16$)  \\
    \bottomrule
  \end{tabular}
\end{table}

\subsubsection{Post-processing using conditional random field (CRF)}
\label{subsub:crf}
Post-processing the output of fully convolutional networks using a CRF often leads to smooth and better aligned mask with the region of interest \citep{ChenPKMY14}. To this end, we use the CRF implementation of \citep{KrahenbuhlK11}\footnote{\href{https://github.com/lucasb-eyer/pydensecrf}{https://github.com/lucasb-eyer/pydensecrf}}. The results are presented in Tab.\ref{tab:tab7-sup-mat}. Following the notation in \citep{KrahenbuhlK11}, we set ${w^{(1)} = w^{(2)} = 1}$. We set, over all the methods, ${\theta_{\alpha}=13, \theta_{\beta}=3, \theta_{\gamma}=3}$ for $2$  iterations, over GlaS, and ${\theta_{\alpha}=19, \theta_{\beta}=11, \theta_{\gamma}=5}$ for $5$  iterations, over CUB5, CUB, and OxF. Tab.\ref{tab:tab7-sup-mat} shows a slight improvement in term of F1$^+$ and slight degradation in term of F1$^-$. When investigating the processed masks, we found that the CRF helps in improving the mask only when the mask covers precisely large part of the region of interest. In this case, the CRF helps spreading the mask over the region. In the case where there is high false positive, or the mask misses largely the region, the CRF does not help. We can see as well that the CRF increases slightly the false positive by spreading the mask out of the region of interest. Since our method has small false positive --i.e., the produced mask covers mostly the region of interest and avoids stepping outside it-- using CRF helps in improving both F1$^+$ and F1$^-$ in most cases.

\begin{table}[h!]
  \caption{Pointwise localization performance of different WSL models over the different test sets when post-processing the predicted masks using CRF \citep{KrahenbuhlK11}. ${*^{+}}$ indicates improvement while ${*^{-}}$ indicates degradation of performance  compared to Tab.\ref{tab:tab3-pl}. Deep MIL \citep{ilse2018attention} is discarded since the produced plans do not form probability over the classes axe at pixel level which is required for the CRF input \citep{KrahenbuhlK11}. To preserve horizontal space, we rename the methods CAM-Avg, CAM-Max, CAM-LSE, Grad-CAM to Avg, Max, LSE, G-C, respectively.}
  \label{tab:tab7-sup-mat}
  \centering
  \small
  \resizebox{1.\linewidth}{!}{
  \begin{tabular}{l|l|l|l|l|l|l|l|l}
    \toprule
    &  \multicolumn{8}{c}{Pixel level} \\
    Method &  \multicolumn{4}{c|}{F1$^+$ (\%)} &  \multicolumn{4}{c}{F1$^-$ (\%)}\\
    \cmidrule{2-9}
                      &  GlaS    &   CUB5 & CUB & OxF    &    GlaS   &  CUB5  & CUB & OxF \\
    \midrule
    Avg \citep{zhou2016learning}          &  $66.90$   & $34.90^{+}$ & $30.86^{-}$  & $57.66^{-}$  & $17.65^{-}$   &  $66.85^{-}$ & $80.63^{-}$ & $65.84^{-}$\\
    Max  \citep{oquab2015object}         &  $66.02^{+}$   & $5.22^{-}$ & $35.59^{+}$  & $41.04^{+}$  & $26.22^{-}$   &  $75.23^{-}$ & $91.14^{+}$ & $73.36^{-}$\\
    LSE  \citep{PinheiroC15cvpr,sun2016pronet}        &  $66.06^{+}$   & $7.85^{+}$ & $36.32^{-}$  & $41.49^{+}$  & $27.75^{-}$   &  $77.09^{-}$ & $91.26^{+}$ & $72.57^{-}$\\
    Wildcat \citep{durand2017wildcat}         &  $67.22^{+}$   & $36.41^{+}$ & $33.03^{-}$  & $54.47^{+}$  & $22.74^{-}$   &  $74.91^{-}$ & $84.97^{+}$ & $72.92^{-}$\\
    G-C \citep{selvaraju2017grad}         &  $66.33^{+}$   & $36.44^{+}$ & $30.86^{-}$  & $57.65^{-}$  & $20.76^{-}$   &  $67.91^{-}$ & $80.63^{-}$ & $65.83^{-}$\\
    \midrule
    Ours ($u=4$)      &  $\bm{72.58}^{+}$   & $\bm{54.69}^{+}$ & $\bm{53.35^{+}}$  & $42.69^{-}$  & $\bm{66.49^{-}}$   &  $\bm{91.06^{+}}$ & $\bm{92.30^{+}}$ & $\bm{75.84^{+}}$\\
    \bottomrule
  \end{tabular}
}
\end{table}

\clearpage
\newpage

\clearpage
\newpage

\clearpage

\end{document}